\documentclass[letterpaper]{article} 
\usepackage{aaai2026}  
\usepackage{times}  
\usepackage{helvet}  
\usepackage{courier}  
\usepackage[hyphens]{url}  
\usepackage{graphicx} 
\def\UrlFont{\rm}  
\usepackage{natbib}  
\usepackage{caption} 
\usepackage{amsmath}
\usepackage{amssymb}
\usepackage{booktabs}

\usepackage{algorithm}
\usepackage{algorithmic}

\usepackage{newfloat}
\usepackage{listings}
\DeclareCaptionStyle{ruled}{labelfont=normalfont,labelsep=colon,strut=off} 
\floatstyle{ruled}
\newfloat{listing}{tb}{lst}{}
\floatname{listing}{Listing}
\title{BrainMCLIP: Brain Image Decoding with Multi-Layer feature Fusion of CLIP}

\author{
    Tian Xia\textsuperscript{\rm 1},
    Zihan Ma\textsuperscript{\rm 1},
    Xinlong Wang\textsuperscript{\rm 1},
    Qing Liu\textsuperscript{\rm 1},
    Xiaowei He\textsuperscript{\rm 1},
    Tianming Liu\textsuperscript{\rm 2},
    Yudan Ren\textsuperscript{\rm 1}
}
\affiliations{
    \textsuperscript{\rm 1}Northwest University, Xian, China\\
    \textsuperscript{\rm 2}University of Georgia, GA, USA\\
    yudan.ren@nwu.edu.cn
}

\usepackage{bibentry}

\begin{document}

\maketitle

\begin{abstract}
Decoding images from fMRI often involves mapping brain activity to CLIP's final semantic layer. 
To capture finer visual details, many approaches add a parameter-intensive VAE-based pipeline.
However, these approaches overlook rich object information within CLIP's intermediate layers and contradicts the brain's functionally hierarchical.
We introduce BrainMCLIP, which pioneers a parameter-efficient, multi-layer fusion approach guided by human visual system's functional hierarchy, eliminating the need for such a separate VAE pathway.
BrainMCLIP aligns fMRI signals from functionally distinct visual areas (low-/high-level) to corresponding intermediate and final CLIP layers, respecting functional hierarchy.
We further introduce a Cross-Reconstruction strategy and a novel multi-granularity loss.
Results show BrainMCLIP achieves highly competitive performance, particularly excelling on high-level semantic metrics where it matches or surpasses SOTA(state-of-the-art) methods, including those using VAE pipelines.
Crucially, it achieves this with substantially fewer parameters, demonstrating a reduction of 71.7\%(Table.\ref{tab:compare_clip_vae}) compared to top VAE-based SOTA methods, by avoiding the VAE pathway.
By leveraging intermediate CLIP features, it effectively captures visual details often missed by CLIP-only approaches, striking a compelling balance between semantic accuracy and detail fidelity without requiring a separate VAE pipeline.
\end{abstract}

\section{Introduction}
\label{sec:Introduction}
Understanding complex brain functions and advancing brain-computer interfaces (BCIs) heavily rely on brain decoding\citep{du2022fmri}.
Functional magnetic resonance imaging (fMRI) offers a non-invasive window into the brain, capturing high-resolution activity patterns particularly valuable for decoding visual perception \citep{allen2022massive}.
Significant recent advancements leverage the power of deep learning, particularly combining CLIP \citep{radford2021learningtransferablevisualmodels} and Diffusion models \citep{ho2020denoising}.
This combination has enabled remarkable progress in reconstructing visual stimuli directly from fMRI signals.
The prevailing approach typically maps fMRI data, often aggregated from the entire visual cortex, to the final layer embeddings of CLIP's vision model and text model to guide the image generation process\citep{linMindReaderReconstructing2022, scottiReconstructingMindsEye2023, ozcelik2023natural, luMindDiffuserControlledImage2023, liuBrainCLIPBridgingBrain2023, scottiMindEye2SharedSubjectModels2024, wangMindBridgeCrosssubjectBrain2024}.
This strategy is common in the field and is based on aligning the semantic processing of the high-level visual cortex with the semantic nature of CLIP's final text and image embedding.

\begin{figure}[tb]
    \begin{center}
        \includegraphics[width=1.0\linewidth]{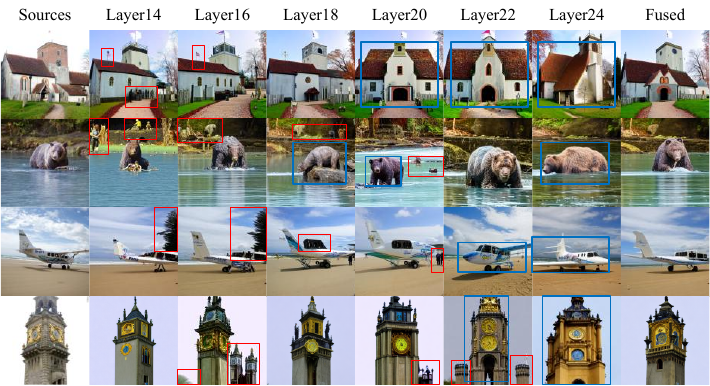}
    \end{center}
    \caption{
    Reconstructions guided by different CLIP vision layers reveal a clear trade-off. Intermediate layers capture fine details but introduce semantic noise (red boxes), while the final layer ensures semantic consistency at the cost of detail accuracy (blue boxes).
    Our proposed fusion of intermediate (layers 10-20) and final layers ('Fused') achieves a compelling balance between detail fidelity and semantic coherence.
    More results are shown in Appendix.A.
    }
    \label{fig:layer_recon_compare}
\end{figure}

Despite these successes, limitations arise particularly from how features of the CLIP vision model are utilized.
Firstly, relying solely on its final layer inherently neglects the rich, fine-grained visual details crucial for faithful reconstruction, as this layer primarily captures semantic information \citep{lanClearCLIPDecomposingCLIP2024, sunCLIPerHierarchicallyImproving2024}.
Recognizing this limitation, many researchers attempt to recover these details by resorting to a separate, parameter-intensive pipeline: training an additional mapping model to project fMRI signals onto the latent space of a Variational Autoencoder(VAE) \citep{ozcelik2023natural, scottiReconstructingMindsEye2023, scottiMindEye2SharedSubjectModels2024}, which introduces substantial parameter overhead and architectural complexity.
Interestingly, while these methods seek external detail features from VAE, computer vision research suggests that rich object detail information is already present in CLIP's own intermediate layers \citep{singhaAPPLeNetVisualAttention2023, liCascadeCLIPCascadedVisionLanguage2024}.
Our preliminary experiment visually confirm this: images reconstructed solely from intermediate layers of CLIP vision model capture finer details compared to final-layer reconstructions, but also tend to introduce semantically irrelevant content or distortions, termed 'noise'(Fig. \ref{fig:layer_recon_compare}). 
Notably, even simple averaging of multi-level features produced compelling reconstructions (Fig. \ref{fig:layer_recon_compare}, 'Fused'), successfully retaining details while preserving semantic coherence and reducing noise.

Furthermore, beyond the specific strategy for detail recovery, a more general limitation persists: current methods typically map fMRI signals from the entire visual cortex uniformly to CLIP's final layer. 
These approaches overlooked the well-established functional hierarchy of the human visual system.
The visual cortex is organized into distinct regions progressing from posterior to anterior areas: early visual cortex (e.g., V1, V2, V3) primarily processes basic features like edges, orientations, and colors \citep{gilbert1983functional}, while higher-level visual areas (e.g., in the ventral stream like LOC, FFA, PPA) integrate these features to represent complex objects, faces, scenes, and semantic categories \citep{tsao2006cortical, rosenke2021probabilistic}.
Such a direct, non-hierarchical mapping strategy disregards this crucial functional hierarchy in human visual cortex and the inherent hierarchical structure of CLIP, hindering optimal feature alignment and reconstruction accuracy.

To overcome these challenges, we introduce BrainMCLIP, a novel framework for fMRI-image reconstruction that is both parameter-efficient and neuro-inspired.
BrainMCLIP directly addresses the limitations by leveraging multi-level representations within the CLIP vision model itself, thereby capturing both semantic content and fine-grained details without resorting to a separate, parameter-costly VAE pipeline.
Crucially, inspired by the functional organization of the visual cortex, BrainMCLIP implements a distinctive mapping strategy: it segregates fMRI data based on functionally specialized visual areas and aligns them with corresponding CLIP layers.
To further refine the mapping and enhance robustness against noise, particularly in intermediate layer features, we incorporate a Cross-Reconstruction strategy.
Additionally, moving beyond standard MSE or contrastive losses that often neglect feature granularity \citep{zhao2016loss, wang2022comprehensive}, we propose a novel multi-granularity loss function based on Centered Kernel Alignment (CKA) and attention map similarity to improve both global and local feature alignment.
Our model focused on a subject-specific setting.

We validated our method on the Natural Scenes Dataset (NSD)\citep{allen2022massive}. Experimental results demonstrate that BrainMCLIP achieves highly competitive decoding accuracy, particularly excelling on high-level semantic metrics while maintaining a compelling balance with detail fidelity, all with significantly fewer parameters compared to VAE-pipeline methods. Our main contributions are:
\begin{itemize}
    \item A novel framework, BrainMCLIP, for parameter-efficient fMRI-based image reconstruction integrating multi-level CLIP features.
    \item A neuro-inspired fMRI data processing and mapping strategy aligned with the functional hierarchy of the human visual cortex and CLIP's layers.
    \item Achieving strong decoding performance, particularly for semantic content, with 71.7\% fewer parameters than leading VAE-based state-of-the-art methods(Table.\ref{tab:compare_clip_vae}).
\end{itemize}

\section{Related works}
\label{sec:Related_works}
\subsection{Brain Image Decoding}
\label{sec:Brain_Image_Decoding}
Early approaches utilized Convolutional Neural Networks (CNNs) and machine learning techniques like linear regression to predict CNN visual features from fMRI data, demonstrating the potential of deep learning in these tasks\cite{horikawa2017generic,shen2019end}.
With the advent of Generative Adversarial Networks (GANs)\cite{goodfellow2020generative}, researchers began mapping fMRI signals to GAN feature spaces.
However, these methods faced challenges in capturing high-level semantic information, leading to images with limited semantic content\cite{shen2019end,shen2019deep}.
Recent breakthroughs have been largely driven by leveraging powerful pre-trained models, namely CLIP \citep{radford2021learningtransferablevisualmodels} for its rich visual-semantic representations and Diffusion Models \citep{ho2020denoising} for their image generation capabilities.
Current methods using this CLIP-Diffusion combined paradigm can be broadly categorized into two main types.
The first, CLIP + VAE Pipeline, maps fMRI signals to CLIP's final layer for semantic guidance but crucially relies on a separate pipeline involving an additional mapping model trained to project fMRI onto a Variational Autoencoder's (VAE) latent space to capture low-level visual features \citep{ozcelik2023natural, scottiReconstructingMindsEye2023, scottiMindEye2SharedSubjectModels2024}.
While achieving strong performance, these approaches introduce significant parameter overhead and architectural complexity, demanding substantial training resources \citep{scottiReconstructingMindsEye2023, scottiMindEye2SharedSubjectModels2024}.
The second type, termed CLIP-Final-Layer Only, simplifies the pipeline by mapping fMRI signals solely to CLIP's final layer embeddings to guide diffusion \citep{liuBrainCLIPBridgingBrain2023, wangMindBridgeCrosssubjectBrain2024}.
This reduces complexity but often struggles to reconstruct fine-grained visual details inherently absent in the final semantic layer.
Our proposed BrainMCLIP offers a novel alternative.
While operating without a VAE like the second type of methods, it significantly departs from them by explicitly leveraging CLIP's intermediate layer features to capture visual details, aiming to achieve the detail fidelity sought by the first type but within a more parameter-efficient framework.

\begin{figure}[t]
    \centering
    \includegraphics[width=0.8\linewidth]{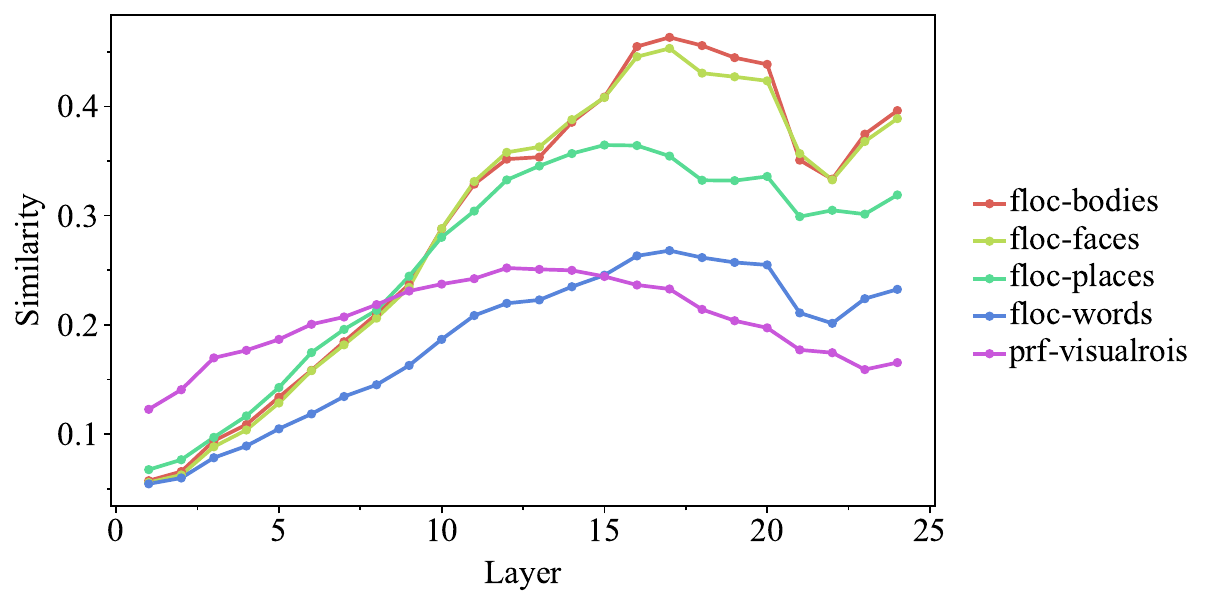}
    \includegraphics[width=0.8\linewidth]{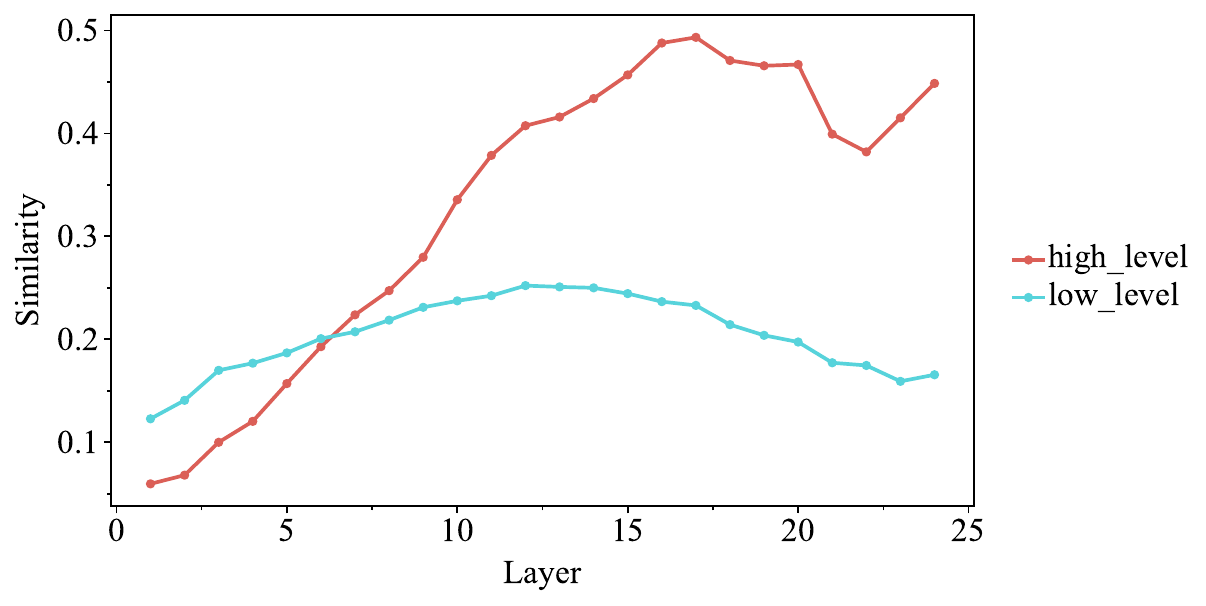}
    
    \caption{
    (Top) Similarity between fMRI features from different visual functional regions (Section. fMRI Data Processing) of subject 01 and CLIP vision model layers. 
    (Bottom) Aggregated high-level visual regions are compared with low-level regions, showing their fMRI feature similarities against CLIP layers. 
    Results for other subjects are provided in Appendix.B.
    }
    \label{fig:rdm_rsa_graph}
\end{figure}

\subsection{Connection of Artificial Neural Networks and Brain Neural Networks}
\label{sec:ANN_BNN}
The design of BrainMCLIP's mapping strategy draws upon converging evidence from neuroscience and computer vision. 
Research reveals significant alignment in feature representations between Artificial Neural Networks (ANNs) and Biological Neural Networks (BNNs) \citep{caucheteux2020language, zhaoCouplingVisualSemantics2022}.
Furthermore, computer vision studies reveal that the CLIP vision model processes visual information in a manner analogous to the human visual cortex: intermediate layers capture fine-grained object details, while later layers encode abstract semantic information \citep{singhaAPPLeNetVisualAttention2023, liCascadeCLIPCascadedVisionLanguage2024, lanClearCLIPDecomposingCLIP2024, sunCLIPerHierarchicallyImproving2024}.

Motivated by this confluence of findings, we propose alignment strategy guided by similarity of both brain and deep model(defined in Sec.fMRI Data Processing) can improve the decoding performance.
We preliminary test this by analyzing the correspondence between multi-level CLIP features and fMRI responses using Representational Similarity Analysis(RSA).
Our analysis for subj01 (Fig.\ref{fig:rdm_rsa_graph}) reveals a hierarchical correspondence: CLIP's intermediate layer features show stronger similarity to fMRI activity in both low-level and high-level visual regions, while its final layer features align more closely with high-level visual areas.
Consistent findings were observed across other subjects, as detailed in Appendix.B.

\section{Methods}
\label{sec:Methods}
\subsection{fMRI Data Processing}
\label{sec:Data_Processing}
Our analysis employed the Natural Scenes Dataset (NSD) \citep{allen2022massive}.
We selected preprocessed fMRI voxels within the ‘NSDGeneral’ region of interest (ROI), which defined by NSD, using the beta maps provided by NSD for each voxel.
This ROI encompasses several subregions within the human visual cortex, including prf-visualrois, responsible for processing basic visual features (e.g. edges, color)\cite{gilbert1983functional}, and the floc-bodies, floc-faces, floc-places, floc-words, which handle more abstract visual information related to object categories\cite{rosenke2021probabilistic}.
Based on these functional distinctions, we classified the prf-visualrois as low-level visual regions and the remaining subregions as high-level visual regions.
For subsequent analysis, we defined fMRI data from all these regions (both low-level and high-level) as \textbf{fMRI-Detail}, denoted as $F_D\in R^{N_D}$, where $N_D$ represents the voxel count in fMRI-Detail. 
The fMRI data from the high-level visual regions is also separately termed \textbf{fMRI-Semantic}, represented as $F_S\in R^{N_S}$, $N_S$ denotes the voxel count in fMRI-Semantic.

\begin{figure*}[t]
    \begin{center}
        \includegraphics[width=0.95\textwidth]{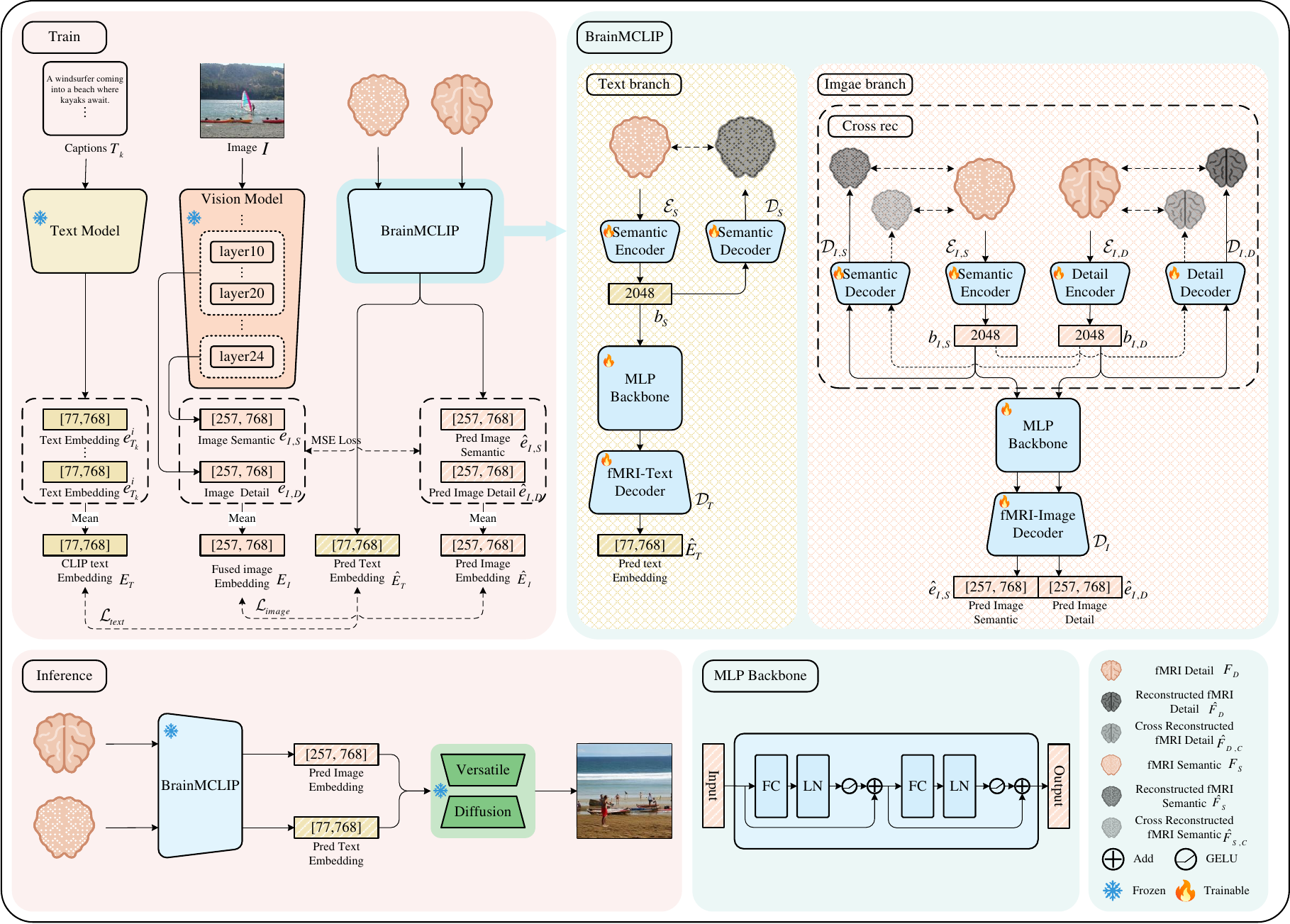}
    \end{center}
    \caption{
    \textbf{Overview of BrainMCLIP.} 
    BrainMCLIP consists of Text and Image branches, both employing an MLP-based backbone and fMRI-to-image decoders. 
    The Text branch aligns fMRI features with the final layer features of the CLIP Text model, while the Image branch aligns them with the fused features from the CLIP Vision model. 
    A Cross-reconstruction module in the Image branch prevents noise learning and improves performance.
    }
    \label{fig:BrainMCLIP_structure}
\end{figure*}

\subsection{BrainMCLIP}
\label{sec:BrainMCLIP}
The sequence of images presented to the subjects is denoted as $\{I^{i}\}_{i=1}^{N}$, where $N$ represents the total number of images.
For each image $I^{i}$, the corresponding set of COCO text descriptions is denoted as $\{T_k^{i}\}_{k=1}^{C^{i}}$, where $C^{i}$ is the number of text descriptions associated with image $I^{i}$.
Each text description $T_k^{i}$ is fed into the CLIP text encoder, and the embedding from its final layer is extracted, denoted as $e_{T_k}^{i}$.
To derive a comprehensive text representation, the text embeddings for each image are averaged, producing the CLIP text embedding $E_T = \frac{1}{C^{i}}\sum_{k=1}^{C^{i}}e_{T_k}^{i}$.
For each image $I^{i}$, the CLIP vision model is employed to extract two distinct feature representations: the \textbf{CLIP-Detail} embedding $e_{I,D}$, derived by averaging features from intermediate layers 11 to 20, and the \textbf{CLIP-Semantic} embedding $e_{I,S}$, obtained from the final layer (layer 24).
(The detailed rationale for selecting these specific vision model layers is provided in Sec.Strategy of middle layer selection). 
To merge the complementary information from these two embeddings, their average is computed, resulting in the \textbf{Fused Image Embedding} $E_{I} = \text{mean}(e_{I,D} + e_{I,S})$, which combines both detailed and semantic information from the vision model and serves as the target representation for the image branch.

\textbf{Overall framework} As illustrated in Fig. \ref{fig:BrainMCLIP_structure}, the BrainMCLIP framework for brain decoding consists of two branches: the Text branch and the Image branch, which are described in detail below.

Grounded in the understanding that the brain's high-level visual cortex is responsible for semantic processing, the text branch is designed to map the fMRI-Semantic signals ($F_S$) to the final layer features of the CLIP text model.
Within the text branch, the fMRI-Semantic $F_S$ is initially fed into the text Semantic Encoder $\mathcal{E}_{S}$, yielding the semantic embedding $b_{S}=\mathcal{E}_{S}\left(F_S\right)$. 
As part of the training objective (detailed in Appendix.C), a corresponding Semantic Decoder $\mathcal{D}_{S}$ is employed to reconstruct the original fMRI-Semantic $\hat{F}_S=\mathcal{D}_{S}\left(b_{S}\right)$, ensuring the encoder captures meaningful semantic features. 
Subsequently, the semantic embedding $b_S$ is processed by a two-layer MLP-based backbone network with residual connections to enhance feature representation. 
The resultant output is then passed to the fMRI-Text Decoder $\mathcal{D}_T$, producing the final predicted CLIP text embedding, denoted as $\hat{E}_T=\mathcal{D}_T\left(MLP\left(b_S\right)\right)$, which is trained to match the ground-truth ${E}_{T}$ derived from the CLIP text model.

Within the image branch, the fMRI-Semantic $F_S$ and the fMRI-Detail $F_D$ are fed into their respective encoders, the Semantic Encoder $\mathcal{E}_{I,S}$ and the Detail Encoder $\mathcal{E}_{I,D}$, yielding the semantic embedding $b_{I,S}=\mathcal{E}_{I,S}\left(F_S\right)$ and the detail embedding $b_{I,D}=\mathcal{E}_{I,D}\left(F_D\right)$. 
Similar to the text branch, and as part of the cross-reconstruction mechanism and loss calculation (Detailed at Appendix.C), these embeddings are also decoded by their corresponding decoders, $\mathcal{D}_{I,S}$ and $\mathcal{D}_{I,D}$, to reconstruct the original fMRI signals $\hat{F}_S=\mathcal{D}_{I,S}\left(b_{I,S}\right)$ and $\hat{F}_D=\mathcal{D}_{I,D}\left(b_{I,D}\right)$, respectively. 
Following the encoders, $b_{I,S}$ and $b_{I,D}$ are processed by a two-layer MLP-based backbone network with residual connections.
The outputs are then passed through the fMRI-Image Decoder $\mathcal{D}_{I}$, yielding the predicted CLIP vision model semantic embedding $\hat{e}_{I,S}=\mathcal{D}_{I}\left(MLP\left(b_{I,S}\right)\right)$ and the predicted CLIP vision model detail embedding $\hat{e}_{I,D}=\mathcal{D}_{I}\left(MLP\left(b_{I,D}\right)\right)$, respectively. 
Finally, these two embeddings are averaged to generate the final Pred Image Embedding $\hat{E}_I = \text{mean}(\hat{e}_{I,S} + \hat{e}_{I,D})$, which is trained to match the fussed ground-truth $E_I$ derived from the CLIP vision model.

During the inference phase, $F_S$ and $F_D$ are input into the BrainMCLIP model to produce $\hat{E}_T$ and $\hat{E}_I$, which are subsequently fed into Versatile Diffusion to generate the final images. More details of the network architecture are presented in Appendix.D.

\textbf{Cross reconstruction mechanism} As observed in our preliminary experiments (Fig. \ref{fig:layer_recon_compare}) and discussed in Sec.Introduction, intermediate layer embeddings from the CLIP vision model can introduce noise, potentially hindering decoding accuracy. 
To mitigate this and enhance the robustness of the learned representations, we propose a cross-reconstruction mechanism within the image branch. 
The core idea is to leverage the semantic information extracted from high-level brain areas ($F_S$) to constrain the feature extraction from the broader detail-focused areas ($F_D$), thereby guiding the model to capture semantically relevant details while suppressing noise.
Specifically, we input the semantic embedding $b_{I,S}$ into the Detail Decoder $\mathcal{D}_{I,D}$ to obtain the cross-reconstructed semantic fMRI signal $\hat{F}_{S,C}=\mathcal{D}_{I,D}(b_{I,S})$. 
Conversely, we input the detail embedding $b_{I,D}$ into the Semantic Decoder $\mathcal{D}_{I,S}$ to obtain the cross-reconstructed detail fMRI signal $\hat{F}_{D,C}=\mathcal{D}_{I,S}(b_{I,D})$. 
The discrepancy between these cross-reconstructions and the original fMRI signals contributes to the cross-reconstruction loss.

\subsection{Multi-Granularity Loss Function}
\label{sec:loss_function}
Prior brain decoding studies often rely on losses like Mean Squared Error (MSE) or contrastive objectives (e.g., InfoNCE, SoftCLIP), which primarily enforce global feature similarity. 
However, these may overlook finer-grained representational differences crucial for detailed reconstruction\citep{scottiReconstructingMindsEye2023, zhao2016loss, wang2022comprehensive}. 
To address this, we propose a Multi-Granularity Loss Function ($\mathcal{L}_{MG}$) that explicitly combines constraints at both global and local (token) levels.

\textit{Global Alignment with CKA ($\mathcal{L}_{CKA}$):} To ensure overall semantic alignment between the predicted embedding $\mathbf{B}$ (e.g., $\hat{E}_T$ or $\hat{E}_I$) and the ground-truth embedding $\mathbf{A}$ (e.g., $E_T$ or $E_I$), we employ the Centered Kernel Alignment (CKA) loss. 
CKA measures the similarity between the representational spaces captured by $\mathbf{A}$ and $\mathbf{B}$. The CKA loss is defined as:
\begin{align}
\mathcal{L}_{CKA}(\mathbf{A}, \mathbf{B}) = 1 - CKA(\mathbf{A}, \mathbf{B})
\end{align}
where $CKA(\mathbf{A}, \mathbf{B})$ is computed based on the Hilbert-Schmidt Independence Criterion (HSIC). (Definitions of CKA and HSIC are provided in Appendix.E).

\textit{Fine-grained Alignment with Cosine Similarity ($\mathcal{L}_{Sims}$):} To capture finer-grained, token-level relationships, particularly the relative importance or focus within the embedding sequence (inspired by attention map, details in Appendix.D), we introduce a similarity loss based on cosine distances. 
Assuming $\mathbf{A}, \mathbf{B} \in \mathbb{R}^{m \times d}$ have the same sequence length ($m$), we consider the first token $\mathbf{t}_{A,1}, \mathbf{t}_{B,1}$ and the remaining tokens $\mathbf{T}_{A,1}, \mathbf{T}_{B,1}$. We compute vectors $\mathbf{s}_A, \mathbf{s}_B$ where each element represents the cosine similarity between the first token and a subsequent token within $\mathbf{A}$ and $\mathbf{B}$, respectively. 
The fine-grained loss encourages these similarity patterns to match:
\begin{align}
\mathcal{L}_{Sims}(\mathbf{A}, \mathbf{B}) = \mathcal{L}_{MSE}(\mathbf{s}_A, \mathbf{s}_B)
\end{align}
This loss component focuses on the internal relational structure of the embeddings, complementing the global CKA alignment.

\textit{Combined Multi-Granularity Loss ($\mathcal{L}_{MG}$):} Our proposed multi-granularity loss is the weighted sum of the global and fine-grained components:
\begin{align}
\mathcal{L}_{MG}(\mathbf{A}, \mathbf{B}) = \mathcal{L}_{CKA}(\mathbf{A}, \mathbf{B}) + \mathcal{L}_{Sims}(\mathbf{A}, \mathbf{B}) \label{eq:lmg_combined}
\end{align}
This combined loss promotes alignment at both coarse and fine representational levels.

\begin{figure}[t]
    \centering
    \includegraphics[width=0.7\columnwidth]{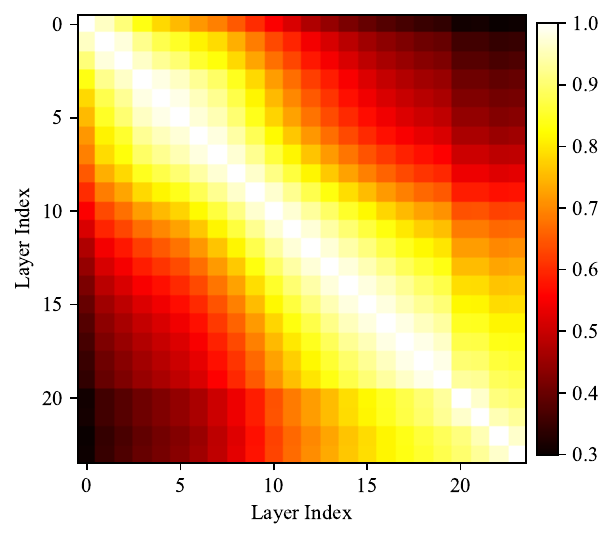}
    \caption{
    The heatmap illustrating the Centered Kernel Alignment (CKA) between intermediate layer features of the CLIP vision model.
    }
    \label{fig:image_cka_map}
\end{figure}

\begin{figure}[h!]
    \begin{center}
        \includegraphics[width=1\linewidth]{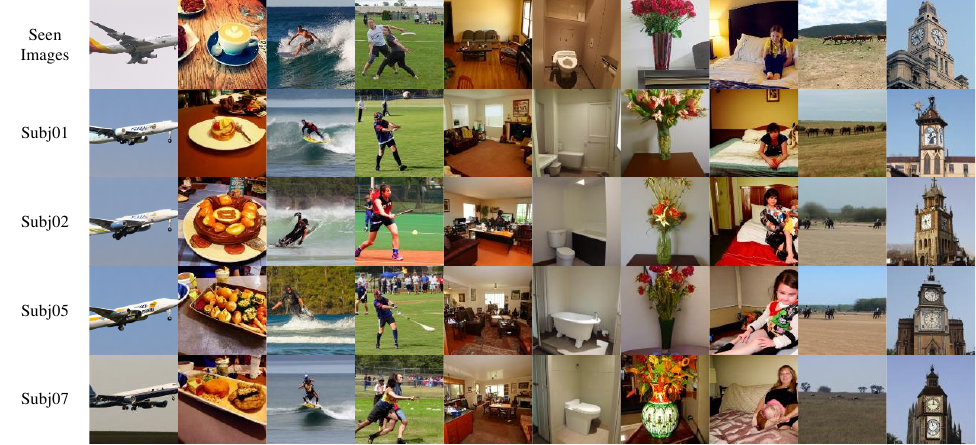}
    \end{center}
    \caption{
    The image reconstruction results of BrainMCLIP.
    We show more results in Appendix.F.
    }
    \label{fig:recon_results}
\end{figure}

\textbf{Total Loss Calculation:}

\textit{Text Branch}: The total loss $\mathcal{L}_{text}$ combines the multi-granularity alignment loss $\mathcal{L}_{MG}(E_T, \hat{E}_T)$ with an MSE loss for fMRI reconstruction $\mathcal{L}_{MSE}(F_S, \hat{F}_S)$:
\begin{align}
\mathcal{L}_{text} = \mathcal{L}_{MG}(E_T, \hat{E}_T) + \mathcal{L}_{MSE}(F_S, \hat{F}_S) \label{eq:loss_text_final}
\end{align}
where $\hat{F}_S = \mathcal{D}_S(\mathcal{E}_S(F_S))$.

\textit{Image Branch}: The total loss $\mathcal{L}_{image}$ includes the multi-granularity alignment loss $\mathcal{L}_{MG}(E_I, \hat{E}_I)$, the cross-reconstruction loss $\mathcal{L}_{Crec}$ (defined in Eq. \ref{eq:crec_loss_in_loss_sec} in Appendix.D), and direct MSE losses on the pre-fusion embeddings $\mathcal{L}_{MSE, I} = \mathcal{L}_{MSE}(e_{I,S}, \hat{e}_{I,S}) + \mathcal{L}_{MSE}(e_{I,D}, \hat{e}_{I,D})$:
\begin{align}
\mathcal{L}_{image} &= \mathcal{L}_{MG}(E_I, \hat{E}_I) + \mathcal{L}_{Crec} + \mathcal{L}_{MSE, I} \label{eq:loss_image_final}
\end{align}

\begin{table*}[htb]
    \caption{
    Performance comparison of BrainMCLIP with methods utilizing CLIP and a VAE pipeline for low-level detail reconstruction.
    Noted that MindEye2 is a cross-subject model.
    Parameter counts (Params) are approximate estimates of the mapping model size.
    \textbf{Bold} indicates best, \underline{underlined} indicates second-best within this comparison group.
    }
    \label{tab:compare_clip_vae}
    \centering
    \setlength{\tabcolsep}{3.5pt}
    \begin{tabular}{@{}cccccccccc@{}}
        \toprule
        \multicolumn{1}{c}{\textbf{Methods}} & \multicolumn{4}{c}{\textbf{Low-Level}} & \multicolumn{4}{c}{\textbf{High-Level}} & \textbf{Params $\downarrow$} \\ 
        \cmidrule(lr){2-5} \cmidrule(lr){6-9}
         & \textbf{PixCorr $\uparrow$} & \textbf{SSIM $\uparrow$} & \textbf{Alex(2) $\uparrow$} & \textbf{Alex(5) $\uparrow$} & \textbf{Incep $\uparrow$} & \textbf{CLIP $\uparrow$} & \textbf{EffNet-B $\downarrow$} & \textbf{Swav $\downarrow$} & \\ \midrule
        Brain-Diffuser & 0.254 & \underline{0.356} & 94.2\% & 96.2\% & 87.2\% & 91.5\% & 0.775 & 0.423 & -- \\
        Mind-Diffuser & -- & 0.354 & -- & -- & -- & 76.5\% & -- & -- & 3B \\
        MindEye & 0.309 & 0.323 & 94.7\% & \underline{97.8\%} & 93.8\% & \underline{94.1\%} & \underline{0.645} & 0.367 & \underline{1.45B} \\
        MindEye2 & \underline{0.322} & \textbf{0.431} & \textbf{96.1\%} & \textbf{98.6\%} & \textbf{95.4\%} & 93.0\% & \textbf{0.619} & \textbf{0.344} & 2.58B \\ 
        \cmidrule(lr){1-10}
        \begin{tabular}[c]{c@{}c@{}}BrainMCLIP(Ours) \\ + Low-Level(MindEye)\end{tabular} 
        & \textbf{0.330} & 0.312 & \underline{94.9\%} & \underline{97.8\%} & \underline{94.5\%} & \textbf{94.7\%} & \underline{0.645} & \underline{0.356} & \textbf{0.73B} \\
        \bottomrule
    \end{tabular}
\end{table*}

\begin{table*}[htb]
    \caption{
    Performance comparison of BrainMCLIP with methods relying solely on CLIP features (without VAE pipeline).
    MindBridge and MindEye2 (wo-VAE) are cross-subject models.
    Parameter counts (Params.) are approximate estimates.
    \textbf{Bold} indicates best, \underline{underlined} indicates second-best.
    }. 
    \label{tab:compare_clip_only}
    \centering
    \setlength{\tabcolsep}{3.5pt}
    \begin{tabular}{@{}cccccccccc@{}}
        \toprule
        \multicolumn{1}{c}{\textbf{Methods}} & \multicolumn{4}{c}{\textbf{Low-Level}} & \multicolumn{4}{c}{\textbf{High-Level}} & \textbf{Params $\downarrow$} \\ 
        \cmidrule(lr){2-5} \cmidrule(lr){6-9}
         & \textbf{PixCorr $\uparrow$} & \textbf{SSIM $\uparrow$} & \textbf{Alex(2) $\uparrow$} & \textbf{Alex(5) $\uparrow$} & \textbf{Incep $\uparrow$} & \textbf{CLIP $\uparrow$} & \textbf{EffNet-B $\downarrow$} & \textbf{Swav $\downarrow$} & \\ \midrule
        BrainClip & -- & -- & -- & -- & 86.7\% &\underline{94.8\%} & -- & -- & -- \\
        MindBridge & 0.151 & 0.263 & 87.7\% & 95.5\% & 92.4\% & 94.7\% & 0.712 & 0.418 & \textbf{0.69B} \\
        MindEye (wo-VAE) & \underline{0.194} & \underline{0.308} & \underline{91.7\%} & \textbf{97.4\%} & \underline{93.6\%} & 94.2\% & \underline{0.645} & \underline{0.369} & 1.25B \\
        MindEye2 (wo-VAE) & 0.155 & \textbf{0.309} & 79.6\% & 88.6\% & 85.3\% & 79.5\% & 0.805 & 0.490 & 2.38B \\ 
        \cmidrule(lr){1-10}
        BrainMCLIP (Ours) & \textbf{0.212} & 0.263 & \textbf{91.8\%} & \underline{97.0\%} & \textbf{94.6\%} & \textbf{95.2\%} & \textbf{0.643} & \textbf{0.354} & \underline{0.73B} \\ 
        \bottomrule
    \end{tabular}
\end{table*}

\subsection{Strategy of middle layer selection}
\label{sec:Strategy_of_middle_layer_selection}
As established in Section \ref{sec:ANN_BNN}, analyses like RSA reveal a hierarchical correspondence between CLIP vision model layers and human visual brain regions (Fig. \ref{fig:rdm_rsa_graph}).
Furthermore, our preliminary reconstructions (Fig. \ref{fig:layer_recon_compare}) demonstrated that while CLIP's intermediate layers are crucial for capturing fine-grained visual details, they can also introduce significant noise. 
These observations highlight the need for a strategy to select the most effective intermediate layers. 
Our strategy is based on a deeper analysis of both the layer-wise alignment with fMRI signals and the internal representational consistency across intermediate layers of CLIP vision model.



We examined the alignment between CLIP layers and fMRI using RSA (Fig. \ref{fig:rdm_rsa_graph}). 
While intermediate layers showed strong similarity to brain activity related to object processing, we observed a notable decrease in similarity for layers 21-23, particularly with higher-level visual areas, suggesting these layers might be less optimal for image reconstruction.
In addition, we analyzed inter-layer feature similarity using CKA (Fig. \ref{fig:image_cka_map}). 
The CKA map revealed a significant shift in representation starting around layer 21, with layers 21-24 exhibiting substantially lower similarity to the earlier intermediate layers compared to layers within the 11-20 range.

RSA and CKA analyzes suggesting suboptimal characteristics for layers 21-24, we conducted extensive evaluations to directly assess their impact on reconstruction performance.
As detailed in Appendix.G, these experiments consistently demonstrated that features from layers 21-23 degraded reconstruction quality, while utilizing layers 10-20 alongside the final layer (layer 24) yielded superior results.
These validations strongly supported our decision to exclude layers 21-23.
Therefore we select layers 10-20 for detail representation and the last layer(layer24) for semantic representation.

\begin{table*}[htb] 
    \renewcommand{\arraystretch}{1.2} 
    \caption{
    Ablation study of the BrainMCLIP model architecture.
    Performance impact of removing key components within the Image branch: the detail pathway (using intermediate layer features), the semantic pathway (using the final layer feature), and the Cross-Reconstruction mechanism.
    Average results across four subjects. 
    \textbf{Bold} denotes the best performance.
    }
    \label{tab:ablation_structure}
    \centering
    \setlength{\tabcolsep}{3.5pt}
    \begin{tabular}{@{}ccccccccc@{}} 
    \toprule
    & \multicolumn{4}{c}{\textbf{Low-Level}} & \multicolumn{4}{c}{\textbf{High-Level}} \\ 
    \cmidrule(lr){2-5} \cmidrule(lr){6-9}
    & \textbf{PixCorr $\uparrow$} & \textbf{SSIM $\uparrow$} & \textbf{Alex(2) $\uparrow$} & \textbf{Alex(5) $\uparrow$} & \textbf{Incep $\uparrow$} & \textbf{CLIP $\uparrow$} & \textbf{EffNet-B $\downarrow$} & \textbf{Swav $\downarrow$} \\ 
    \midrule
    Text Branch Only
    & 0.065 & 0.107 & 59.67\% & 73.56\% & 78.41\% & 78.46\% & 0.858 & 0.555 \\
    Text + Image Semantic
    & 0.088 & 0.211 & 74.16\% & 87.21\% & 90.03\% & 90.38\% & 0.725 & 0.452 \\ 
    Text + Image Detail
    & 0.166 & 0.259 & 86.64\% & 93.80\% & 92.91\% & 93.95\% & 0.668 & 0.386 \\
    \begin{tabular}[c]{@{}c@{}}Text + Image Semantic \\ + Image Detail\end{tabular} 
    & 0.204 & 0.257 & 90.90\% & 96.55\% & 94.03\% & 93.93\% & 0.654 & 0.363 \\
    \midrule
    \begin{tabular}[c]{@{}c@{}c@{}}Text + Image Semantic \\ + Image Detail \\ + Cross Reconstruction(Ours)\end{tabular}
    & \textbf{0.212} & \textbf{0.263} & \textbf{91.8\%} & \textbf{97.0\%} & \textbf{94.6\%} & \textbf{95.2\%} & \textbf{0.643} & \textbf{0.354} \\
    \bottomrule
    \end{tabular}
\end{table*}

\begin{table*}[htb] 
    \caption{
    Ablation of the multi-granularity loss functions. 
    Compares the full proposed loss ('Ours') against standard baselines (MSE + Contrastive losses) and ablations of its components (MSE + Sims, MSE + CKA).
    \textbf{Bold} denotes the best performance.
    }
    \label{tab:ablation_loss}
    \centering
    \setlength{\tabcolsep}{3.5pt}
    \begin{tabular}{@{}ccccccccc@{}} 
    \toprule
    & \multicolumn{4}{c}{\textbf{Low-Level}} & \multicolumn{4}{c}{\textbf{High-Level}} \\ 
    \cmidrule(lr){2-5} \cmidrule(lr){6-9}
    & \textbf{PixCorr $\uparrow$} & \textbf{SSIM $\uparrow$} & \textbf{Alex(2) $\uparrow$} & \textbf{Alex(5) $\uparrow$} & \textbf{Incep $\uparrow$} & \textbf{CLIP $\uparrow$} & \textbf{EffNet-B $\downarrow$} & \textbf{Swav $\downarrow$} \\ 
    \midrule
    MSE + InfoNCE                                                                      
    & 0.205 & 0.239 & 90.10\% & 96.57\% & 93.06\% & 93.27\% & 0.669 & 0.371 \\
    MSE + SoftCLIP
    & 0.201 & 0.238 & 89.84\% & 96.63\% & 93.12\% & 93.11\% & 0.657 & 0.374 \\
    \midrule
    MSE + Sims
    & 0.197 & 0.214 & 89.68\% & 95.82\% & 92.23\% & 92.22\% & 0.693 & 0.370 \\
    MSE + CKA
    & 0.200 & 0.233 & 89.06\% & 95.99\% & 93.11\% & 93.07\% & 0.676 & 0.376 \\
    \midrule
    MSE + CKA + Sims(Ours) 
    & \textbf{0.212} & \textbf{0.263} & \textbf{91.8\%} & \textbf{97.0\%} & \textbf{94.6\%} & \textbf{95.2\%} & \textbf{0.643} & \textbf{0.354} \\
    \bottomrule
    \end{tabular}
\end{table*}

\section{Results and Analysis}
\label{sec:Results_and_Analysis}
Fig.\ref{fig:recon_results} shows our decoding examples.
Quantitative evaluation used low-level (PixCorr, SSIM\citep{wang2004image}, AlexNet(2), AlexNet(5)\citep{krizhevsky2012imagenet}) and high-level (Inception\citep{szegedy2016rethinking}, CLIP, EffNet-B\citep{tan2019efficientnet}, SwAV\citep{caron2020unsupervised}) metrics.
We also report model parameter counts (Params).
Results are averaged across four subjects.
We compared BrainMCLIP against six SOTA methods: Brain-Diffuser\citep{ozcelik2023natural}, Mind-Diffuser\citep{luMindDiffuserControlledImage2023}, MindEye\citep{scottiReconstructingMindsEye2023}, MindEye2\citep{scottiMindEye2SharedSubjectModels2024}, BrainCLIP \citep{liuBrainCLIPBridgingBrain2023}, and MindBridge\citep{wangMindBridgeCrosssubjectBrain2024}.
These methods differ significantly in their approach to capturing low-level visual details and overall architecture.
To provide a clear comparison, we present the results in two separate tables, comparing BrainMCLIP against methods employing a VAE pipeline (Table.\ref{tab:compare_clip_vae}) and those relying solely on CLIP features (Table.\ref{tab:compare_clip_only}).
For a fair comparison, we utilize the low-level pipeline from MindEye for image generation with our model's outputs in Table.\ref{tab:compare_clip_vae}.
Notably, while MindEye and MindEye2 are capable of operating without a VAE, their non-VAE configurations in Table.\ref{tab:compare_clip_only}) were included in the CLIP only group for this comparison.


While ranking just behind MindEye2 on several metrics in Table. \ref{tab:compare_clip_vae}, BrainMCLIP achieves this competitive performance with a remarkable 71.7\% reduction in parameters, highlighting a superior trade-off between accuracy and efficiency.
The results suggest BrainMCLIP effectively leveraging CLIP's own features for semantics and details in a parameter-efficient manner, demonstrating the value of exploring multi-level CLIP features as an alternative to VAEs for detail recovery. (Note: MindEye2 is cross-subject).

When compared to methods that solely rely on CLIP on Table.\ref{tab:compare_clip_only}, BrainMCLIP consistently outperforms most existing approaches across both low-level and high-level metrics. 
This superior performance can be attributed to BrainMCLIP’s effective utilization of object detail information embedded within the intermediate layers of the CLIP vision model. 
This not only allows for improved low-level feature reconstruction but also facilitates the accurate reconstruction of high-level semantic information. 
These findings underscore BrainMCLIP’s ability to achieve a robust balance between reconstructing low-level details and capturing high-level semantic features.

We further projected the model's output embeddings back into the fMRI space, which confirmed that the predicted semantic and detail features maintained the expected distinct correlations with high-level and low-level visual areas, respectively, reinforcing our model's alignment with the brain's functional hierarchy (Appendix.H).

\section{Ablations}
\label{sec:Ablations}
Our ablation results are obtained by averaging the performance across four subjects.

\textbf{Structure Ablation} 
We evaluated key architectural components by systematically ablating parts of the Image branch, which includes Semantic, Detail, and Cross-Reconstruction modules (Table \ref{tab:ablation_structure}). 
The full model('Ours'), integrating all components, achieved the best overall performance. 
Removing the Cross-Reconstruction module degraded performance ("Text + Image Semantic + Image Detail" row), particularly for high-level metrics, underscoring its crucial role in enhancing robustness and potentially mitigating noise.
Ablating the Image Detail pathway("Text + Image Semantic" row) caused a significant drop across all metrics, confirming that intermediate features ($e_{I,D}$) are vital for reconstruction fidelity.
Similarly, removing the Image Semantic pathway while keeping the detail pathway ("Text + Image Detail" row) impaired performance compared to using both pathways, especially on semantic metrics, highlighting the importance of the final layer ($e_{I,S}$) for semantic guidance.
Finally, using only the Text branch ("Text Branch Only" row) yielded the worst performance across all metrics, establishing a baseline and confirming the need for our dual-branch design.

\textbf{Losses Ablation} 
We evaluated our Multi-Granularity Loss ($\mathcal{L}_{MG}$), combining MSE, global (CKA) and fine-grained (Sims) alignment, against standard baselines("MSE + InfoNCE" row and "MSE + SoftCLIP" row) and its own ablations.
Our loss surpassed standard baselines("MSE + InfoNCE", "MSE + SoftCLIP"), particularly in SSIM and high-level metrics.
This suggests standard contrastive losses may not optimally balance feature levels.
Using only Sims loss (MSE+Sims) yielded poor results, likely by overemphasizing local relations.
Using only CKA loss (MSE + CKA) performed better, but was still inferior to the full loss.
Our full loss (Ours), achieved the best results across most metrics, demonstrating that combining blobal (CKA) and local (Sims) constraints creates a more balanced representation.

\section{Conclusion}
\label{sec:Conclusion}
We introduced BrainMCLIP, a parameter-efficient framework for fMRI-based image reconstruction leveraging multi-level CLIP vision features via a neuro-inspired mapping. 
By aligning fMRI from distinct visual areas with corresponding CLIP layers, BrainMCLIP achieves a strong balance between semantic accuracy and detail fidelity without parameter-intensive VAE pipelines. 
Our work underscores the potential of integrating multi-level CLIP features with brain-functional principles for advance neural decoding.

\bibliography{BrainMCLIP}


\setlength{\leftmargini}{20pt}
\makeatletter\def\@listi{\leftmargin\leftmargini \topsep .5em \parsep .5em \itemsep .5em}
\def\@listii{\leftmargin\leftmarginii \labelwidth\leftmarginii \advance\labelwidth-\labelsep \topsep .4em \parsep .4em \itemsep .4em}
\def\@listiii{\leftmargin\leftmarginiii \labelwidth\leftmarginiii \advance\labelwidth-\labelsep \topsep .4em \parsep .4em \itemsep .4em}\makeatother

\setcounter{secnumdepth}{0}
\renewcommand\thesubsection{\arabic{subsection}}
\renewcommand\labelenumi{\thesubsection.\arabic{enumi}}

\newcounter{checksubsection}
\newcounter{checkitem}[checksubsection]

\newcommand{\checksubsection}[1]{%
  \refstepcounter{checksubsection}%
  \paragraph{\arabic{checksubsection}. #1}%
  \setcounter{checkitem}{0}%
}

\newcommand{\checkitem}{%
  \refstepcounter{checkitem}%
  \item[\arabic{checksubsection}.\arabic{checkitem}.]%
}
\newcommand{\question}[2]{\normalcolor\checkitem #1 #2 \color{blue}}
\newcommand{\ifyespoints}[1]{\makebox[0pt][l]{\hspace{-15pt}\normalcolor #1}}

\section*{Reproducibility Checklist}









\checksubsection{General Paper Structure}
\begin{itemize}

\question{Includes a conceptual outline and/or pseudocode description of AI methods introduced}{(yes/partial/no/NA)}
yes

\question{Clearly delineates statements that are opinions, hypothesis, and speculation from objective facts and results}{(yes/no)}
yes

\question{Provides well-marked pedagogical references for less-familiar readers to gain background necessary to replicate the paper}{(yes/no)}
yes

\end{itemize}
\checksubsection{Theoretical Contributions}
\begin{itemize}

\question{Does this paper make theoretical contributions?}{(yes/no)}
no

	\ifyespoints{\vspace{1.2em}If yes, please address the following points:}
        \begin{itemize}
	
	\question{All assumptions and restrictions are stated clearly and formally}{(yes/partial/no)}
	Type your response here

	\question{All novel claims are stated formally (e.g., in theorem statements)}{(yes/partial/no)}
	Type your response here

	\question{Proofs of all novel claims are included}{(yes/partial/no)}
	Type your response here

	\question{Proof sketches or intuitions are given for complex and/or novel results}{(yes/partial/no)}
	Type your response here

	\question{Appropriate citations to theoretical tools used are given}{(yes/partial/no)}
	Type your response here

	\question{All theoretical claims are demonstrated empirically to hold}{(yes/partial/no/NA)}
	Type your response here

	\question{All experimental code used to eliminate or disprove claims is included}{(yes/no/NA)}
	Type your response here
	
	\end{itemize}
\end{itemize}

\checksubsection{Dataset Usage}
\begin{itemize}

\question{Does this paper rely on one or more datasets?}{(yes/no)}
yes

\ifyespoints{If yes, please address the following points:}
\begin{itemize}

	\question{A motivation is given for why the experiments are conducted on the selected datasets}{(yes/partial/no/NA)}
	yes

	\question{All novel datasets introduced in this paper are included in a data appendix}{(yes/partial/no/NA)}
	NA

	\question{All novel datasets introduced in this paper will be made publicly available upon publication of the paper with a license that allows free usage for research purposes}{(yes/partial/no/NA)}
	NA

	\question{All datasets drawn from the existing literature (potentially including authors' own previously published work) are accompanied by appropriate citations}{(yes/no/NA)}
	yes

	\question{All datasets drawn from the existing literature (potentially including authors' own previously published work) are publicly available}{(yes/partial/no/NA)}
	yes

	\question{All datasets that are not publicly available are described in detail, with explanation why publicly available alternatives are not scientifically satisficing}{(yes/partial/no/NA)}
	yes

\end{itemize}
\end{itemize}

\checksubsection{Computational Experiments}
\begin{itemize}

\question{Does this paper include computational experiments?}{(yes/no)}
yes

\ifyespoints{If yes, please address the following points:}
\begin{itemize}

	\question{This paper states the number and range of values tried per (hyper-) parameter during development of the paper, along with the criterion used for selecting the final parameter setting}{(yes/partial/no/NA)}
	partial

	\question{Any code required for pre-processing data is included in the appendix}{(yes/partial/no)}
	no

	\question{All source code required for conducting and analyzing the experiments is included in a code appendix}{(yes/partial/no)}
	no

	\question{All source code required for conducting and analyzing the experiments will be made publicly available upon publication of the paper with a license that allows free usage for research purposes}{(yes/partial/no)}
	yes
        
	\question{All source code implementing new methods have comments detailing the implementation, with references to the paper where each step comes from}{(yes/partial/no)}
	yes

	\question{If an algorithm depends on randomness, then the method used for setting seeds is described in a way sufficient to allow replication of results}{(yes/partial/no/NA)}
	yes

	\question{This paper specifies the computing infrastructure used for running experiments (hardware and software), including GPU/CPU models; amount of memory; operating system; names and versions of relevant software libraries and frameworks}{(yes/partial/no)}
	yes

	\question{This paper formally describes evaluation metrics used and explains the motivation for choosing these metrics}{(yes/partial/no)}
	yes

	\question{This paper states the number of algorithm runs used to compute each reported result}{(yes/no)}
	yes

	\question{Analysis of experiments goes beyond single-dimensional summaries of performance (e.g., average; median) to include measures of variation, confidence, or other distributional information}{(yes/no)}
	no

	\question{The significance of any improvement or decrease in performance is judged using appropriate statistical tests (e.g., Wilcoxon signed-rank)}{(yes/partial/no)}
	no

	\question{This paper lists all final (hyper-)parameters used for each model/algorithm in the paper’s experiments}{(yes/partial/no/NA)}
	partial

\end{itemize}
\end{itemize}

\end{document}


\appendix

\section{More results about intermediate layer reconstruction of CLIP vision model}
\label{sec:more_intermediate_recon_results}
\begin{figure}[H] 
    \centering
    \includegraphics[width=1\linewidth]{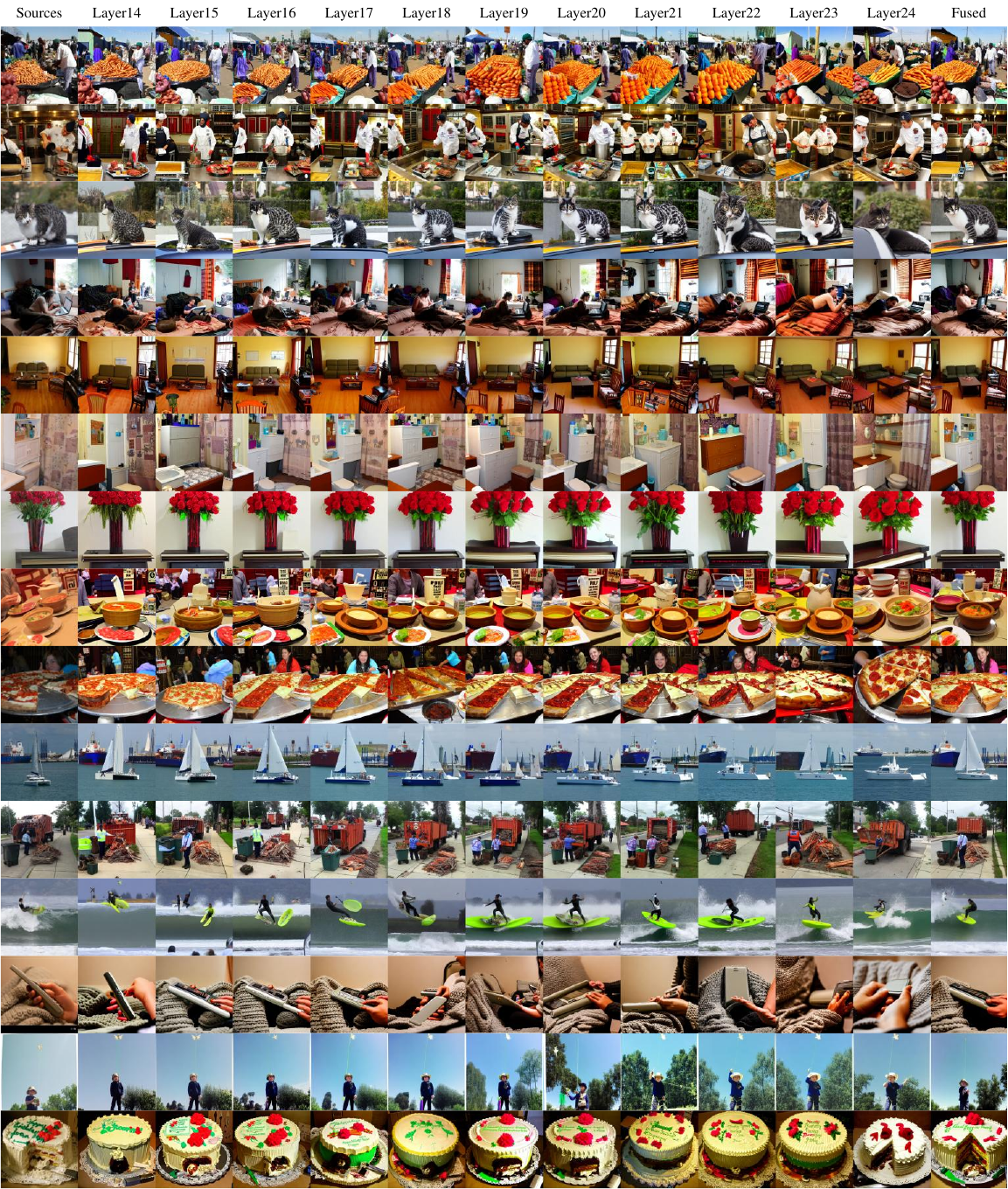}
    \caption{
    Additional examples illustrating image reconstructions using Versatile Diffusion guided by different CLIP vision model layers.
    }
    \label{fig:recon_results_more}
\end{figure}

\section{Similarity between fMRI features of visual functional regions and CLIP vision model layers for the ramaining subjects}
\label{sec:rdm_rsa_other_subjs}
\begin{figure}[htbp]
    \centering
    \begin{subfigure}{0.3\textwidth}
        \centering
        \includegraphics[width=\textwidth]{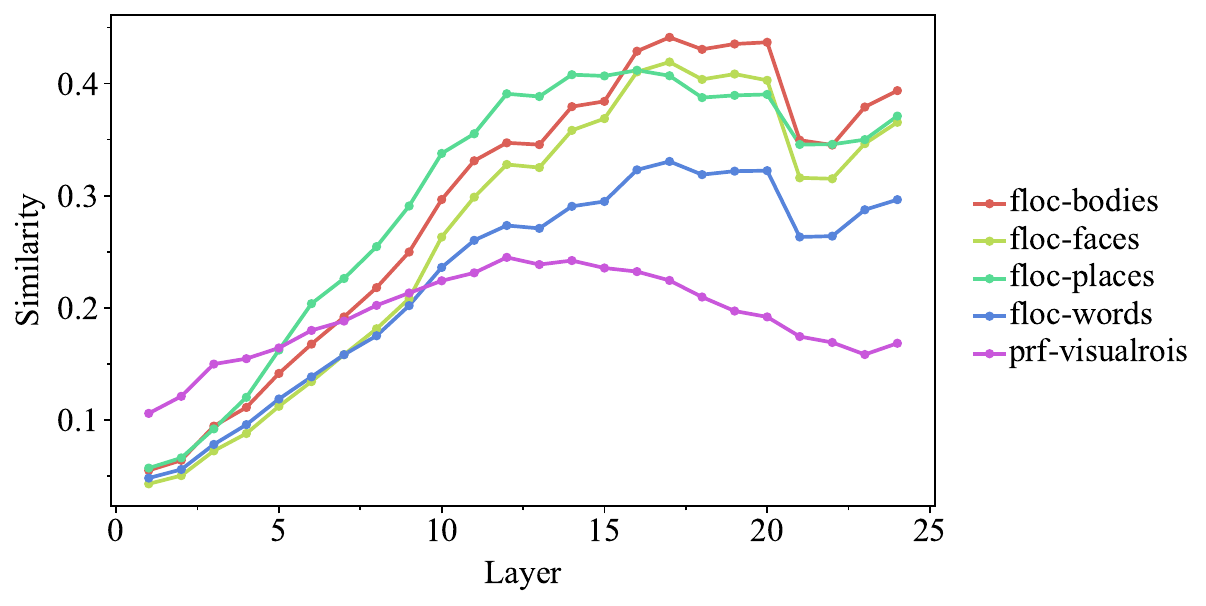}
        \includegraphics[width=\textwidth]{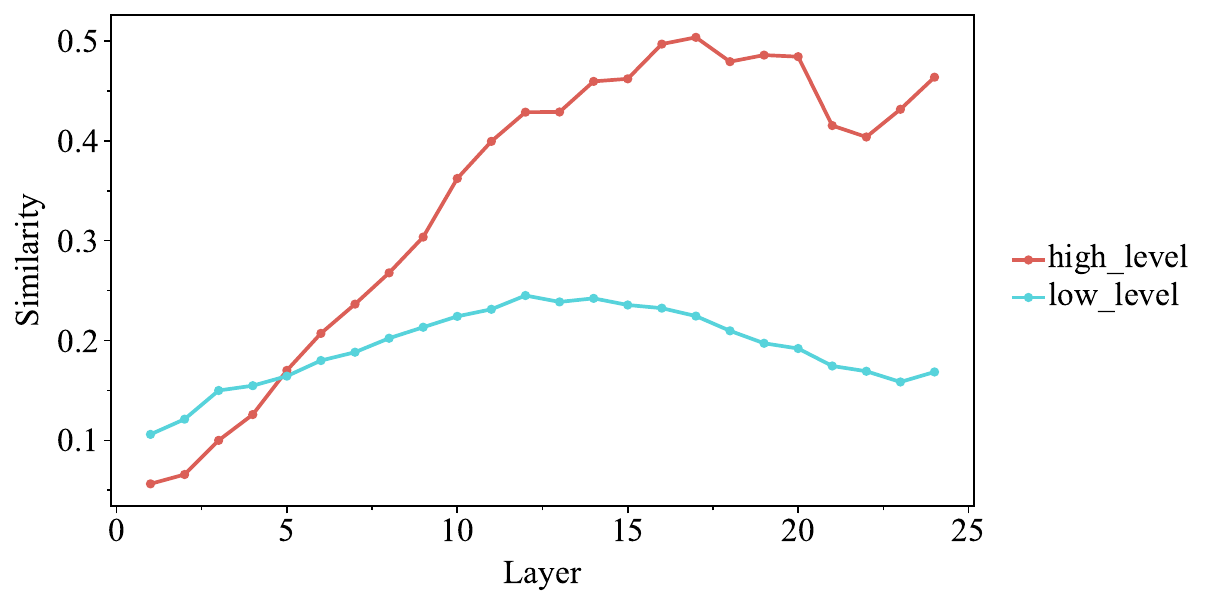}
        \captionsetup{labelformat=empty}
        \caption{subj02} 
    \end{subfigure}
    \begin{subfigure}{0.3\textwidth} 
        \centering 
        \includegraphics[width=\textwidth]{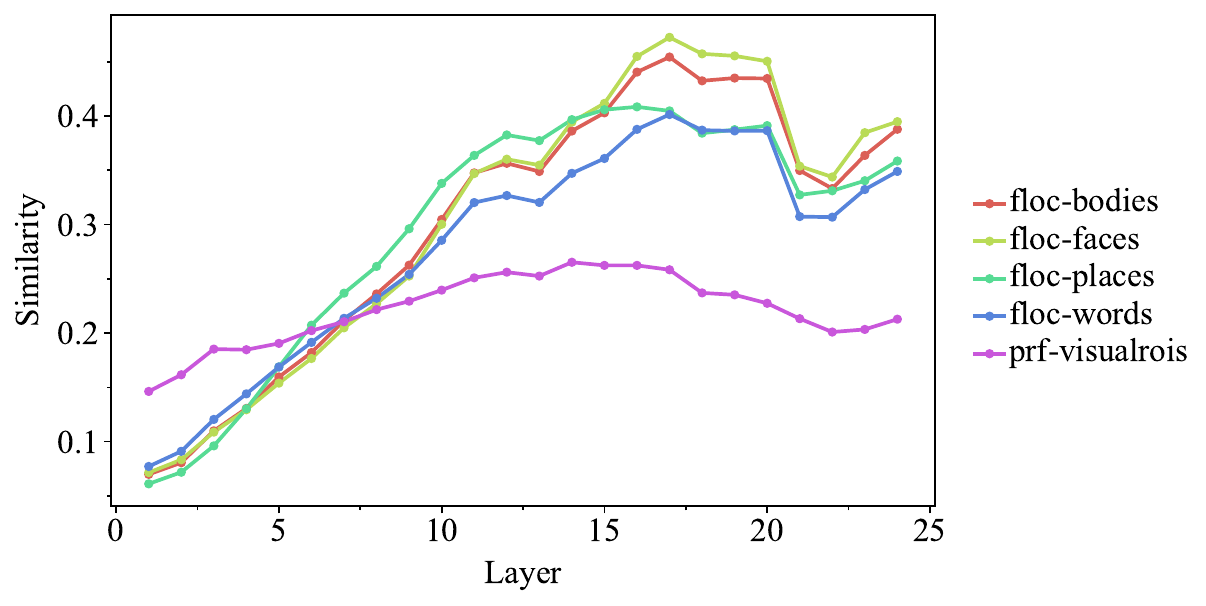}
        \includegraphics[width=\textwidth]{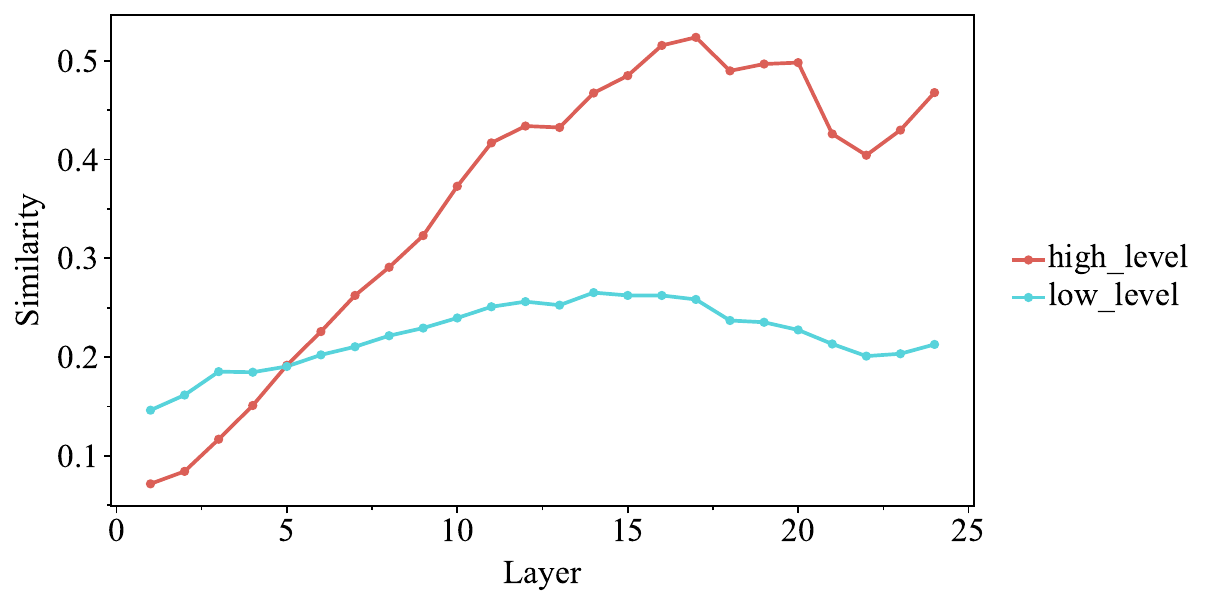}
        \captionsetup{labelformat=empty}
        \caption{subj05}
    \end{subfigure}
    \begin{subfigure}{0.3\textwidth} 
        \centering
        \includegraphics[width=\textwidth]{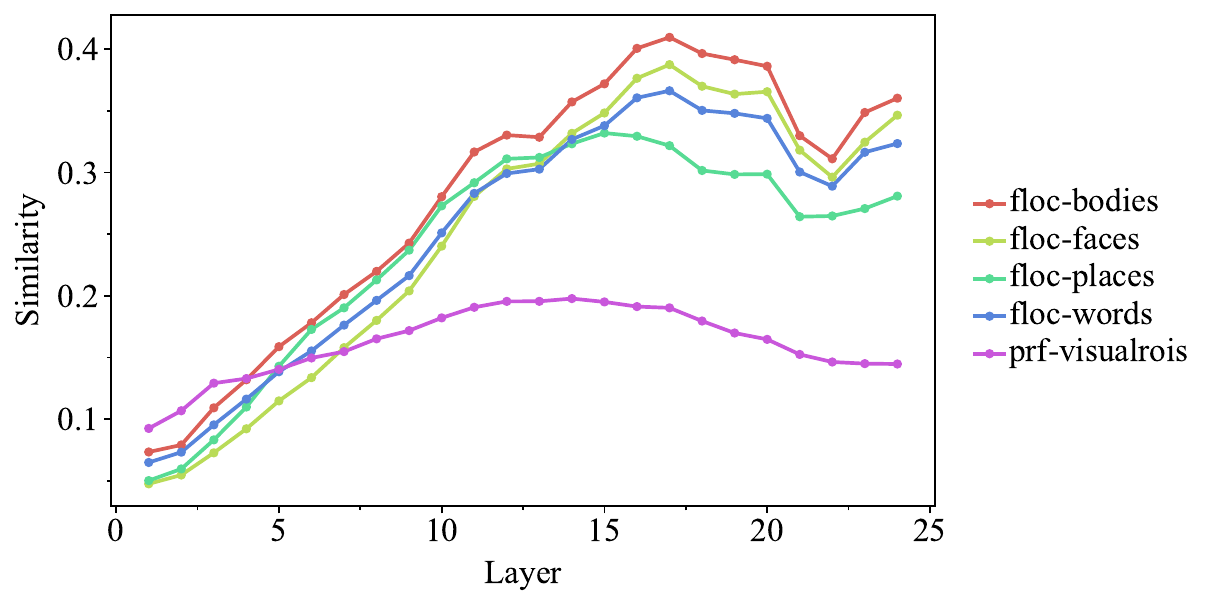}
        \includegraphics[width=\textwidth]{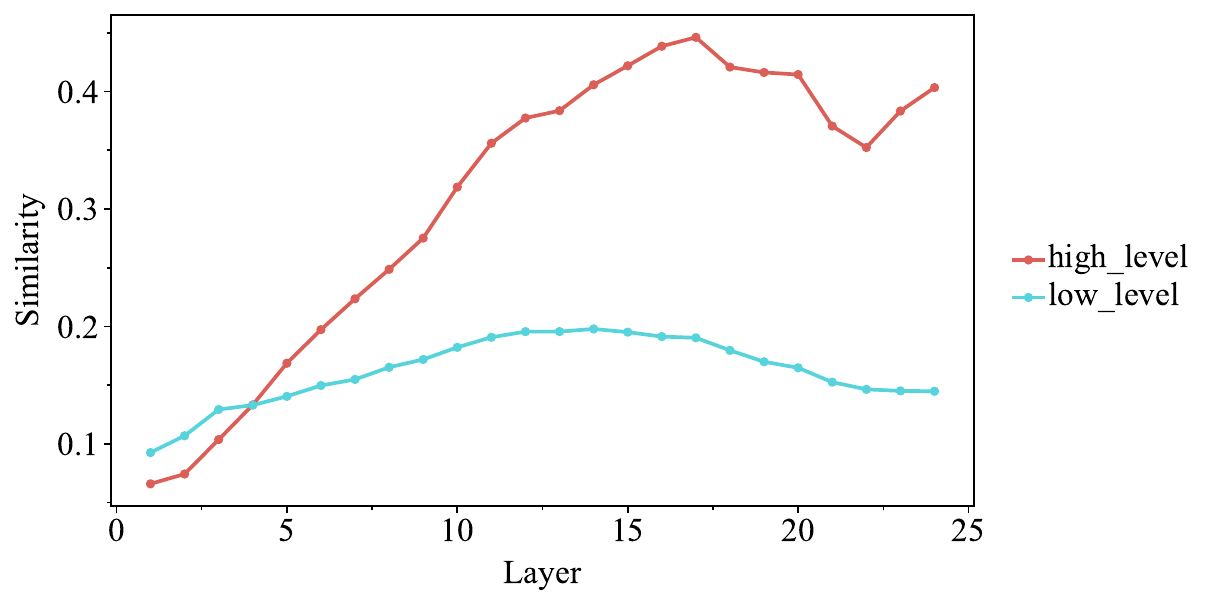} 
        \captionsetup{labelformat=empty}
        \caption{subj07} 
    \end{subfigure} 
    
    \caption{
    Representational Similarity Analysis (RSA) results comparing fMRI features from different human visual functional regions with CLIP vision model layers for the remaining subjects (subj02, subj05, and subj07).
    For each subject, the upper plot shows similarity curves for fine-grained functional regions, while the lower plot aggregates these into high-level and low-level visual regions.
    These results are consistent with the findings for subj01 (Fig. \ref{fig:rdm_rsa_graph}), showing a hierarchical correspondence and a notable similarity decrease in later intermediate layers.
    }
    \label{fig:rdm_rsa_other_subjs}
\end{figure}

To demonstrate the consistency of the fMRI-CLIP layer correspondence across participants, we present the Representational Similarity Analysis (RSA) results for the remaining three subjects (subj02, subj05, and subj07) in Figure \ref{fig:rdm_rsa_other_subjs}. 
As shown in the figure, the similarity trends for these subjects closely mirror those observed for subj01 (Fig. \ref{fig:rdm_rsa_graph}). 
Specifically, the intermediate layers of the CLIP vision model consistently show higher similarity to fMRI activity in areas processing object-level information, while the final layers align more strongly with high-level semantic areas. 
Furthermore, the decrease in similarity between layers 21-23 and fMRI features is reproducible across these subjects. 
These consistent findings support the generalizability of our observations regarding the hierarchical alignment between human visual cortex and CLIP vision model layers, reinforcing the rationale for our multi-level feature selection strategy.

\section{Multi-Granularity Loss Function}
\label{sec:loss_function}
To effectively guide the model training, we propose a comprehensive loss function incorporating constraints at multiple levels of granularity. This includes a multi-granularity loss ($\mathcal{L}_{MG}$) targeting the alignment between predicted and ground-truth CLIP embeddings, and reconstruction losses ensuring the encoders capture relevant fMRI information.

\textbf{Global Alignment with CKA} Given two embedding matrices $\mathbf{A} \in \mathbb{R}^{m \times d}$ (representing brain signals) and $\mathbf{B} \in \mathbb{R}^{n \times d}$ (representing CLIP model embeddings), we compute the CKA similarity to measure their global alignment.
CKA is derived from the Hilbert-Schmidt Independence Criterion (HSIC) and is defined as:
\begin{align}
CKA\left(\mathbf{A},\mathbf{B}\right)=\frac{HSIC\left(\mathbf{A},\mathbf{B}\right)}{\sqrt{HSIC\left(\mathbf{A},\mathbf{A}\right)}\sqrt{HSIC\left(\mathbf{B},\mathbf{B}\right)}}
\end{align}
where HSIC is calculated as:
\begin{align}
\mathrm{HSIC}(\mathbf{A},\mathbf{B})=\frac{1}{(m-1)^2}\mathrm{tr}(\mathbf{K}\mathbf{H}\mathbf{L}\mathbf{H})
\end{align}
with $\mathbf{K} \in \mathbb{R}^{m \times m}$ and $\mathbf{L} \in \mathbb{R}^{n \times n}$ being the kernel matrices of $\mathbf{A}$ and $\mathbf{B}$, respectively, and $\mathbf{H}$ the $m \times m$ centering matrix.
The CKA loss is then defined as 
\begin{align}
\mathcal{L}_{CKA}=1-CKA\left(\mathbf{A},\mathbf{B}\right)
\end{align}

\textbf{Fine-grained Alignment with Cosine Similarity} To capture finer-grained information, we introduce a token-level loss.
Assuming $\mathbf{A}$ and $\mathbf{B}$ have the same number of tokens ($m=n$), we denote their first tokens as $\mathbf{t}_{A,1}$ and $\mathbf{t}_{B,1}$, and the remaining tokens as matrices $\mathbf{T}_{A,1}$ and $\mathbf{T}_{B,1}$, which are $\mathbf{A}$ and $\mathbf{B}$ with the first token removed.
We compute cosine similarity vectors $\mathbf{s}_{A}$ (between $\mathbf{t}_{A,1}$ and each token in $\mathbf{T}_{A,1}$) and $\mathbf{s}_{B}$ (between $\mathbf{t}_{A,1}$ and each token in $\mathbf{T}_{B,1}$).
In Transformer models, attention maps are visualizations or matrices that represent the relationships given by one token to another within a sequence. 
The cosine similarity between the class token (first token) and subsequent tokens in a sequence can serve as a proxy for this kind of relationship pattern. 
By aligning the cosine similarity vectors $\mathbf{s}_A$ and $\mathbf{s}_B$, we aim to encourage the predicted embedding ($\mathbf{B}$) to exhibit a similar internal focus and relational structure relative to its first token as the ground-truth embedding ($\mathbf{A}$), analogous to aligning their attention patterns. 
This helps align the internal focus and local dependencies within the embeddings.
The fine-grained loss is the mean squared error (MSE) between $\mathbf{s}_A$ and $\mathbf{s}_B$:
\begin{align}
\mathcal{L}_{Sims}\left(\mathbf{A},\mathbf{B}\right)=MSE\left(\mathbf{s}_A,\mathbf{s}_B\right)
\end{align}

\textbf{Multi-Granularity Loss ($\mathcal{L}_{MG}$)} The core multi-granularity loss combines the global CKA loss and the fine-grained similarity loss:
\begin{align}
\mathcal{L}_{MG}\left(\mathbf{A},\mathbf{B}\right)=\mathcal{L}_{CKA}\left(\mathbf{A},\mathbf{B}\right)+\mathcal{L}_{Sims}\left(\mathbf{A},\mathbf{B}\right)
\end{align}
This component promotes effective alignment between the predicted embeddings ($\hat{E}_T, \hat{E}_I$) and their corresponding ground-truth CLIP embeddings ($E_T, E_I$) at both global and token levels.

\textbf{Total Loss Calculation for Branches}

\textit{Text Branch}: The total loss for the text branch combines the multi-granularity alignment loss with an fMRI reconstruction loss:
\begin{align}
\mathcal{L}_{MG,T} &= \mathcal{L}_{MG}\left(E_T, \hat{E}_T\right) \\
\mathcal{L}_{rec,T} &= \mathcal{L}_{MSE}\left(F_S, \hat{F}_S\right) \\
\mathcal{L}_{text} &= \mathcal{L}_{MG,T} + \mathcal{L}_{rec,T} \label{eq:loss_text}
\end{align}
where $\hat{F}_S$ is the reconstructed fMRI-Semantic signal from the text branch decoder $\mathcal{D}_S$. $\mathcal{L}_{rec,T}$ ensures the text semantic encoder $\mathcal{E}_S$ captures meaningful information from $F_S$.

\textit{Image Branch}: The image branch loss includes the multi-granularity alignment loss, the cross-reconstruction loss, and additional MSE terms constraining the individual predicted CLIP embeddings.
First, we define the Cross-Reconstruction Loss ($\mathcal{L}_{Crec}$), which incorporates errors from both direct fMRI reconstructions ($\hat{F}_S = \mathcal{D}_{I,S}(b_{I,S})$ and $\hat{F}_D = \mathcal{D}_{I,D}(b_{I,D})$) and cross-reconstructions ($\hat{F}_{S,C} = \mathcal{D}_{I,D}(b_{I,S})$ and $\hat{F}_{D,C} = \mathcal{D}_{I,S}(b_{I,D})$) as described in Sec. \ref{sec:BrainMCLIP}:
\begin{align}
    \label{eq:crec_loss_in_loss_sec} 
    \mathcal{L}_{Crec} &= \mathcal{L}_{MSE}\left(F_S, \hat{F}_{S}\right) + \mathcal{L}_{MSE}\left(F_S, \hat{F}_{S,C}\right) \nonumber \\
    &+ \mathcal{L}_{MSE}\left(F_D, \hat{F}_D\right) + \mathcal{L}_{MSE}\left(F_D, \hat{F}_{D,C}\right)
\end{align}
This loss enhances the robustness of the image branch encoders ($\mathcal{E}_{I,S}, \mathcal{E}_{I,D}$).
The multi-granularity loss for the image branch targets the fused embeddings:
\begin{align}
\mathcal{L}_{MG,I} = \mathcal{L}_{MG}\left(E_I, \hat{E}_I\right)
\end{align}
Additionally, we apply MSE losses to the individual predicted CLIP embeddings before fusion:
\begin{align}
\mathcal{L}_{MSE, I} = \mathcal{L}_{MSE}\left(e_{I,S}, \hat{e}_{I,S}\right) + \mathcal{L}_{MSE}\left(e_{I,D}, \hat{e}_{I,D}\right)
\end{align}
This provides direct supervision for the semantic and detail embedding predictions.
The total image branch loss is then the weighted sum of these components:
\begin{align}
\mathcal{L}_{image} &= \mathcal{L}_{MG,I} + \mathcal{L}_{Crec} + \mathcal{L}_{MSE, I} \label{eq:loss_image}
\end{align}

\section{More Details of BrainMCLIP}
\label{sec:Detail_of_BrainMCLIP}
The intermediate layer embeddings of the CLIP Vision model, with dimensions of [257, 1024], are passed through its final fully connected layer to produce embeddings of dimensions [257, 768]. 
In the BrainMCLIP framework, all the encoders and decoders are designed as single-layer Multi-Layer Perceptrons (MLPs) enhanced with residual connections and utilize the Rectified Linear Unit (ReLU) as the activation function. 
To prevent overfitting, dropout is applied. The encoders and decoders each use a dropout rate of 0.15, while the MLP backbone of BrainMCLIP employs a higher dropout rate of 0.5.
Image reconstruction is performed using the Versatile Diffusion model, configured with a text-image ratio of 0.5 and a guidance scale of 5.0. 
For each test sample, eight images are generated concurrently, and the one with the highest CLIP similarity to the stimulus image is selected as the final output. 
We use OpenAI’s CLIP VIT-L/14 checkpoint, with embedding dimensions of 77x768 and 257x768 for the Text and Image branches, respectively. 
Training is performed on a single NVIDIA RTX 4090 GPU, using batch sizes of 250 and 150 for the Text and Image branches, respectively.

\section{CKA and HSIC Definitions}
\label{sec:cka_hsic_details}
The Centered Kernel Alignment (CKA) similarity between two embedding matrices $\mathbf{A} \in \mathbb{R}^{m \times d}$ and $\mathbf{B} \in \mathbb{R}^{n \times d}$ is calculated as:
\begin{align}
CKA\left(\mathbf{A},\mathbf{B}\right)=\frac{HSIC\left(\mathbf{A},\mathbf{B}\right)}{\sqrt{HSIC\left(\mathbf{A},\mathbf{A}\right)}\sqrt{HSIC\left(\mathbf{B},\mathbf{B}\right)}}
\end{align}
where HSIC is the Hilbert-Schmidt Independence Criterion. Given kernel matrices $\mathbf{K} \in \mathbb{R}^{m \times m}$ (derived from $\mathbf{A}$) and $\mathbf{L} \in \mathbb{R}^{n \times n}$ (derived from $\mathbf{B}$), and the $m \times m$ centering matrix $\mathbf{H} = \mathbf{I} - \frac{1}{m}\mathbf{1}\mathbf{1}^T$, HSIC is calculated as:
\begin{align}
\mathrm{HSIC}(\mathbf{A},\mathbf{B})=\frac{1}{(m-1)^2}\mathrm{tr}(\mathbf{K}\mathbf{H}\mathbf{L}\mathbf{H})
\end{align}

\section{More reconstruction results}
\label{sec:more_recon_results}
\begin{figure}[H]
    \begin{center}
        \includegraphics[width=0.95\linewidth]{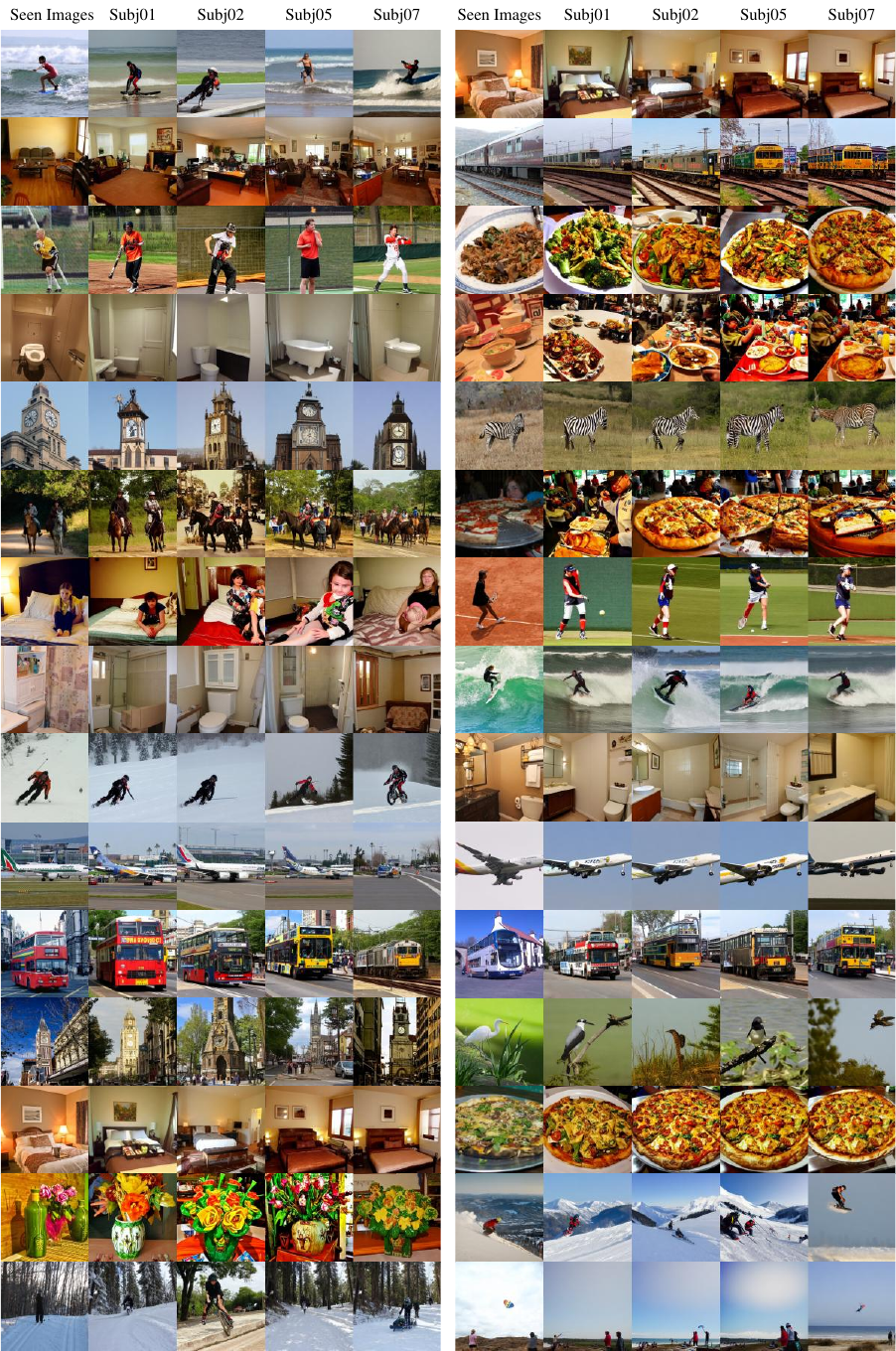}
    \end{center}
    \caption{
    Additional examples of image reconstructions for all four subjects.
    }
    \label{fig:recon_results_more}
\end{figure}

\section{The effect of reconstructing images by the middle layer of CLIP Vision model}
\label{sec:layer_combination_performance}
To empirically validate our selection of CLIP vision layers (10-20 and 24), complementing the RSA/CKA analyses, we performed targeted reconstruction experiments using Versatile Diffusion (Tables.\ref{tab:all_single_middle_layer}-\ref{tab:layer_combination_performance_with_fmri}).
Evaluation of individual intermediate layers (Table.\ref{tab:layer_combination_performance}) revealed a performance drop starting from layer 21, confirming potential issues with layers 21-23 identified earlier.
To further investigate the optimal layer set, we tested various layer combinations fused via averaging (Table~\ref{tab:layer_combination_performance}).
These experiments confirmed that combinations incorporating features from layers 21-23 generally underperformed those that excluded these layers.
More specifically, when comparing strategies for handling layers beyond 20, we found that simply excluding layers 21-23 while retaining the final semantic layer (layer 24) -- i.e., using a combination like (10-20) with (24) -- yielded significantly better results than excluding all layers from 21 through 24 (e.g., using only (10-20) without (24)).
Among all tested configurations, the (10-20) combined with (24) consistently produced the best overall reconstruction performance in these experiments.
Crucially, we also confirmed this finding within the full BrainMCLIP framework (Table~\ref{tab:layer_combination_performance_with_fmri}). Training the model with different layer sets using fMRI data unequivocally showed that using layers 10-20 and 24 achieved the highest decoding accuracy.
These consistent results across different evaluation settings provide robust empirical support for excluding layers 21-23 while retaining layer 24, leading to our chosen (10-20), (24) configuration for BrainMCLIP.

\begin{table}[H]
  \centering
  \caption{
  Performance metrics for image reconstruction using features from individual CLIP vision model layers (8-24) by Versatile Diffusion.
  \textbf{Highlighted rows} indicate layers with potential anomalous metric decreases compared to adjacent layers.
  }
  \label{tab:all_single_middle_layer}
  \resizebox{\linewidth}{!}{
    \begin{tabular}{ccccccccc}
      \toprule
      \multirow{2}{*}{\textbf{Layers}} & \multicolumn{4}{c}{\textbf{Low-Level}} & \multicolumn{4}{c}{\textbf{High-Level}} \\
      \cmidrule(lr){2-5} \cmidrule(lr){6-9}
      & PixCorr$\uparrow$ & SSIM$\uparrow$ & Alex(2)$\uparrow$ & Alex(5)$\uparrow$ & Incep$\uparrow$ & CLIP$\uparrow$ & EffNet-B$\downarrow$ & SwAV$\downarrow$ \\
      \midrule
      8  & 0.2706 & 0.2519 & 93.93\% & 94.82\% & 84.56\% & 81.69\% & 0.8231 & 0.5460 \\
      9  & 0.3043 & 0.2795 & 97.34\% & 98.61\% & 92.40\% & 89.62\% & 0.7165 & 0.4421 \\
      10 & 0.3287 & 0.3007 & 98.52\% & 99.45\% & 96.50\% & 93.58\% & 0.6590 & 0.3978 \\
      11 & 0.3490 & 0.3108 & 99.11\% & 99.85\% & 98.58\% & 96.58\% & 0.5741 & 0.3362 \\
      12 & 0.3699 & 0.3100 & 99.38\% & 99.94\% & 99.50\% & 98.40\% & 0.5197 & 0.2903 \\
      13 & 0.3818 & 0.3198 & 99.35\% & 99.96\% & 99.59\% & 99.14\% & 0.4753 & 0.2637 \\
      14 & 0.3833 & 0.3060 & 99.54\% & 99.94\% & 99.73\% & 99.39\% & 0.4550 & 0.2532 \\
      15 & 0.3768 & 0.2986 & 99.47\% & 99.96\% & 99.79\% & 99.69\% & 0.4387 & 0.2412 \\
      16 & 0.3559 & 0.2867 & 99.46\% & 99.94\% & 99.74\% & 99.84\% & 0.4298 & 0.2346 \\
      17 & 0.3520 & 0.2957 & 99.18\% & 99.90\% & 99.72\% & 99.89\% & 0.4243 & 0.2320 \\
      18 & 0.3443 & 0.2971 & 99.01\% & 99.88\% & 99.76\% & 99.89\% & 0.4223 & 0.2287 \\
      19 & 0.3324 & 0.2968 & 98.84\% & 99.84\% & 99.76\% & 99.95\% & 0.4236 & 0.2323 \\
      20 & 0.3233 & 0.2842 & 98.59\% & 99.84\% & 99.76\% & 99.96\% & 0.4222 & 0.2297 \\
      \rowcolor{lightgray} 
      21 & 0.2968 & 0.2512 & 97.97\% & 99.77\% & 99.72\% & 99.89\% & 0.4289 & 0.2347 \\
      \rowcolor{lightgray} 
      22 & 0.2926 & 0.2588 & 97.81\% & 97.81\% & 99.62\% & 99.93\% & 0.4285 & 0.2297 \\
      \rowcolor{lightgray} 
      23 & 0.2943 & 0.2638 & 98.09\% & 99.76\% & 99.66\% & 99.92\% & 0.4370 & 0.2375 \\
      24 & 0.3048 & 0.2826 & 98.30\% & 99.82\% & 99.64\% & 99.96\% & 0.4101 & 0.2201 \\
      \bottomrule
    \end{tabular}
  }
\end{table}

\begin{table}[h]
    \centering
    \caption{
    Performance analysis of image reconstruction via Versatile Diffusion using different combinations of intermediate layers from the CLIP vision model and the final layer of the CLIP text encoder.
    \textbf{Bold} indicates the best performance; \underline{underlined} indicates the second-best.
    }
    \label{tab:layer_combination_performance}
    \resizebox{\linewidth}{!}{
        \begin{tabular}{@{}ccccccccc@{}}
        \toprule
        \multirow{2}{*}{\textbf{Layer Combination}} & \multicolumn{4}{c}{\textbf{Low-Level}} & \multicolumn{4}{c}{\textbf{High-Level}} \\
        \cmidrule(lr){2-5} \cmidrule(lr){6-9}
        & \textbf{PixCorr $\uparrow$} & \textbf{SSIM $\uparrow$} & \textbf{Alex(2) $\uparrow$} & \textbf{Alex(5) $\uparrow$} & \textbf{Incep $\uparrow$} & \textbf{CLIP $\uparrow$} & \textbf{EffNet-B $\downarrow$} & \textbf{Swav $\downarrow$} \\
        \midrule
        Only last layer(24)
        & 0.3048 & 0.2826 & 98.30\% & 99.82\% & 99.64\% & 99.96\% & 0.4101 & 0.2201 \\
        \midrule
        (5-20)          & 0.3788          & 0.2999          & 99.62\%          & 99.96\%          & 99.61\%          & 99.35\%          & 0.4823          & 0.2709          \\
        (5-20), (24)    & 0.4028          & 0.3098          & \underline{99.78\%} & \underline{99.97\%} & 99.80\% & \underline{99.97\%} & 0.3768          & 0.1976          \\
        (5-24)          & 0.3905          & 0.3012          & 99.72\%          & \textbf{99.98\%} & 99.79\% & 99.83\%          & 0.4346          & 0.2368          \\
        (7-20)          & 0.3963          & 0.3121          & 99.63\%          & \textbf{99.98\%} & 99.77\% & 99.70\%          & 0.4388          & 0.2439          \\
        (7-20), (24)    & 0.4060          & 0.3145          & 99.74\%          & \underline{99.97\%} & \underline{99.83\%} & \underline{99.97\%} & 0.3675          & 0.1936          \\
        (7-24)          & 0.4025          & 0.3105          & 99.66\%          & \textbf{99.98\%} & \underline{99.79\%} & 99.85\%          & 0.4099          & 0.2216          \\
        (9-20)          & 0.4016          & \textbf{0.3195} & 99.77\%          & 99.96\%          & 99.78\% & 99.78\%          & 0.4196          & 0.2295          \\
        (9-20), (24)    & \textbf{0.4033} & 0.3180          & \textbf{99.80\%} & \textbf{99.98\%} & 99.81\%          & \textbf{99.98\%} & 0.3669          & 0.1922          \\
        (9-24)          & 0.4052          & 0.3163          & 99.75\%          & \textbf{99.98\%} & 99.82\% & 99.92\%          & 0.3943          & 0.2110          \\
        (10-20)         & 0.4000          & \underline{0.3189} & 99.73\%       & \textbf{99.98\%} & \underline{99.83\%} & 99.72\%          & 0.4110          & 0.2244          \\
        (10-20), (24)   & 0.4021          & 0.3165          & \textbf{99.80\%} & \textbf{99.98\%} & \underline{99.83\%} & \textbf{99.98\%} & \textbf{0.3657} & \textbf{0.1900} \\
        (10-24)         & \underline{0.4032} & 0.3151          & 99.72\%          & \underline{99.97\%} & \underline{99.80\%} & 99.94\%          & 0.3874          & 0.2071          \\
        (11-20)         & \textbf{0.4033} & 0.3153          & 99.64\%          & \textbf{99.98\%} & \underline{99.83\%} & 99.82\%          & 0.4024          & 0.2194          \\
        (11-20), (24)   & 0.3963          & 0.3136          & 99.68\%          & 99.96\%          & 99.82\% & \textbf{99.98\%} & \underline{0.3666} & \underline{0.1913} \\
        (11-24)         & 0.4005          & 0.3113          & 99.74\%          & \textbf{99.98\%} & 99.80\% & 99.95\%          & 0.3865          & 0.2057          \\
        (13-20)         & 0.3896          & 0.3077          & 99.53\%          & 99.96\%          & \textbf{99.84\%} & 99.92\%          & 0.4036          & 0.2167          \\
        (13-20), (24)   & 0.3814          & 0.3061          & 99.61\%          & 99.96\%          & 99.80\% & \textbf{99.98\%} & 0.3676          & 0.1928          \\
        (13-24)         & 0.3843          & 0.3024          & 99.59\%          & \underline{99.97\%} & \textbf{99.84\%} & \underline{99.97\%} & 0.3873          & 0.2056          \\
        (15-20)         & 0.3657          & 0.2992          & 99.37\%          & 99.95\%          & 99.81\%          & 99.92\%          & 0.4060          & 0.2187          \\
        (15-20), (24)   & 0.3699          & 0.3015          & 99.43\%          & 99.96\%          & 99.81\%          & \textbf{99.98\%} & 0.3731          & 0.1951          \\
        (15-24)         & 0.3658          & 0.2947          & 99.30\%          & 99.95\%          & 99.79\% & \underline{99.97\%} & 0.3910          & 0.2082          \\
        \bottomrule
        \end{tabular}%
    }
\end{table}

\begin{table}[H]
    \centering
    \caption{
    Decoding performance within the BrainMCLIP framework using fMRI data (subj01) for different CLIP vision layer combinations.
    This table validates the layer selection strategy by evaluating reconstructions after full model training.
    \textbf{Bold} indicates the best performance per metric; 
    \underline{underlined} indicates the second-best. 
             }
    \label{tab:layer_combination_performance_with_fmri}
    \resizebox{\linewidth}{!}{
        \begin{tabular}{@{}ccccccccc@{}} 
        \toprule
        \multirow{2}{*}{\textbf{Layer Combination}} & \multicolumn{4}{c}{\textbf{Low-Level}} & \multicolumn{4}{c}{\textbf{High-Level}} \\
        \cmidrule(lr){2-5} \cmidrule(lr){6-9}
        & \textbf{PixCorr$\uparrow$} & \textbf{SSIM$\uparrow$} & \textbf{Alex(2)$\uparrow$} & \textbf{Alex(5)$\uparrow$} & \textbf{Incep$\uparrow$} & \textbf{CLIP$\uparrow$} & \textbf{EffNet-B$\downarrow$} & \textbf{SwAV$\downarrow$} \\
        \midrule
        (5-20),(24)             & \textbf{0.226} & 0.254          & 90.4\%          & 96.3\%          & 94.0\%          & 93.7\%          & 0.668          & 0.374          \\
        (7-20),(24)             & \textbf{0.226} & 0.253          & \underline{90.6\%} & 96.5\%          & 93.9\%          & 93.4\%          & 0.659          & 0.370          \\
        (9-20),(24)             & \underline{0.225} & \underline{0.260} & 90.4\%          & 96.0\%          & 93.8\%          & 93.7\%          & 0.656          & 0.366          \\
        (10-20),(24) & 0.222          & \textbf{0.265} & \textbf{92.5\%} & \textbf{97.6\%} & \textbf{94.4\%} & \textbf{95.1\%} & \textbf{0.647} & \textbf{0.353} \\
        (11-20),(24)            & 0.214          & 0.250          & 91.0\%          & \underline{96.8\%} & 94.1\%          & 94.4\%          & 0.652          & \underline{0.359} \\
        (13-20),(24)            & 0.185          & 0.214          & 87.3\%          & 95.6\%          & 92.8\%          & 93.6\%          & 0.667          & 0.374          \\
        (15-20),(24)            & 0.199          & 0.240          & 89.0\%          & 95.9\%          & \underline{94.7\%} & \underline{94.7\%} & \underline{0.648} & 0.364          \\
        (17-20),(24)            & 0.194          & 0.250          & 87.9\%          & 95.1\%          & 94.0\%          & 94.5\%          & 0.651          & 0.370          \\
        \bottomrule
        \end{tabular}%
    }
\end{table}

\section{Validation of Predicted Feature Alignment}
\label{sec:validation_of_predicted_feature_alignment}

\begin{figure*}[h!]
    \centering
    \begin{minipage}{0.32\textwidth}
        \centering
        \includegraphics[width=\linewidth]{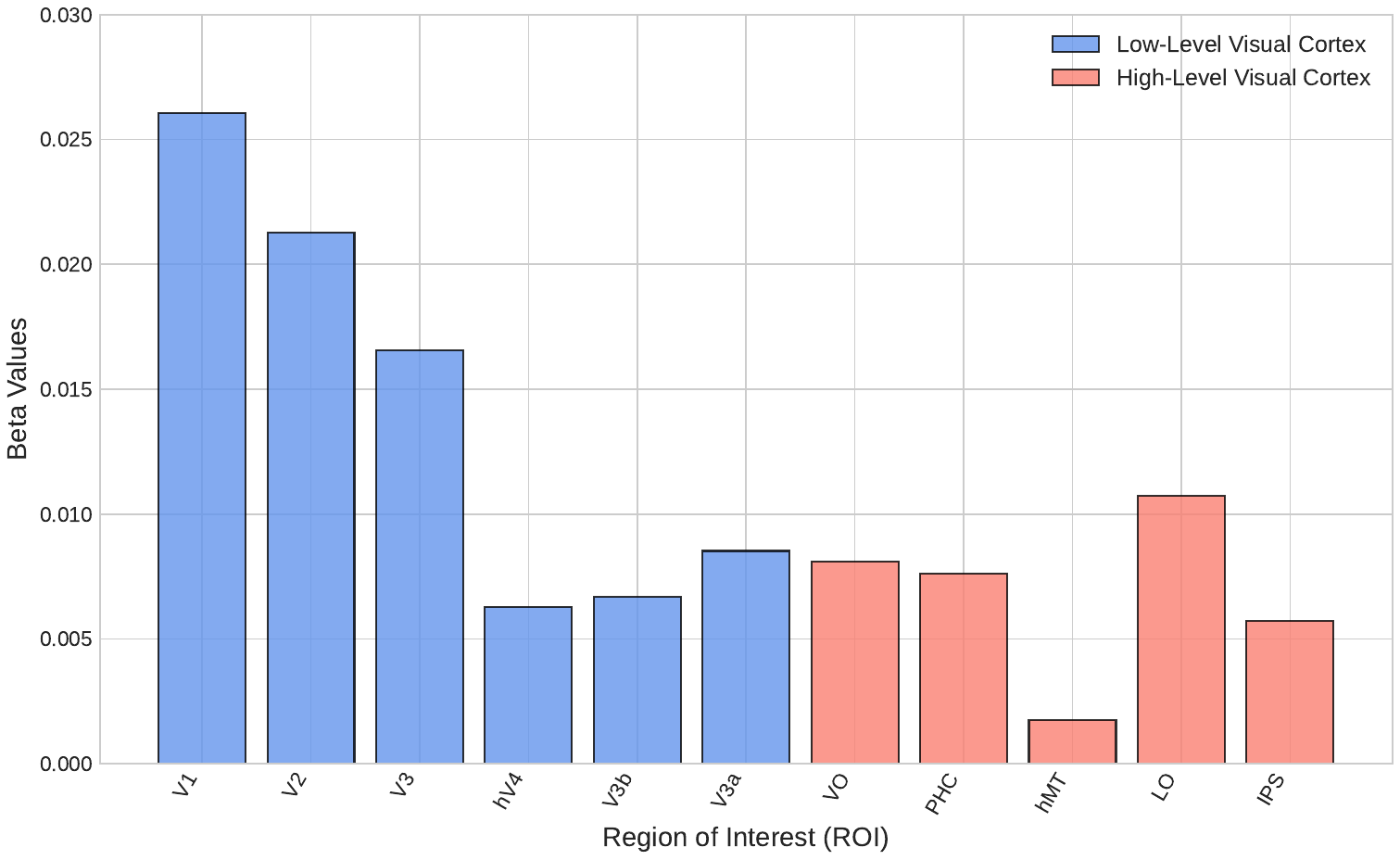}
        \small (A) Image Detail Beta Value
    \end{minipage}
    \hfill
    \begin{minipage}{0.32\textwidth}
        \centering
        \includegraphics[width=\linewidth]{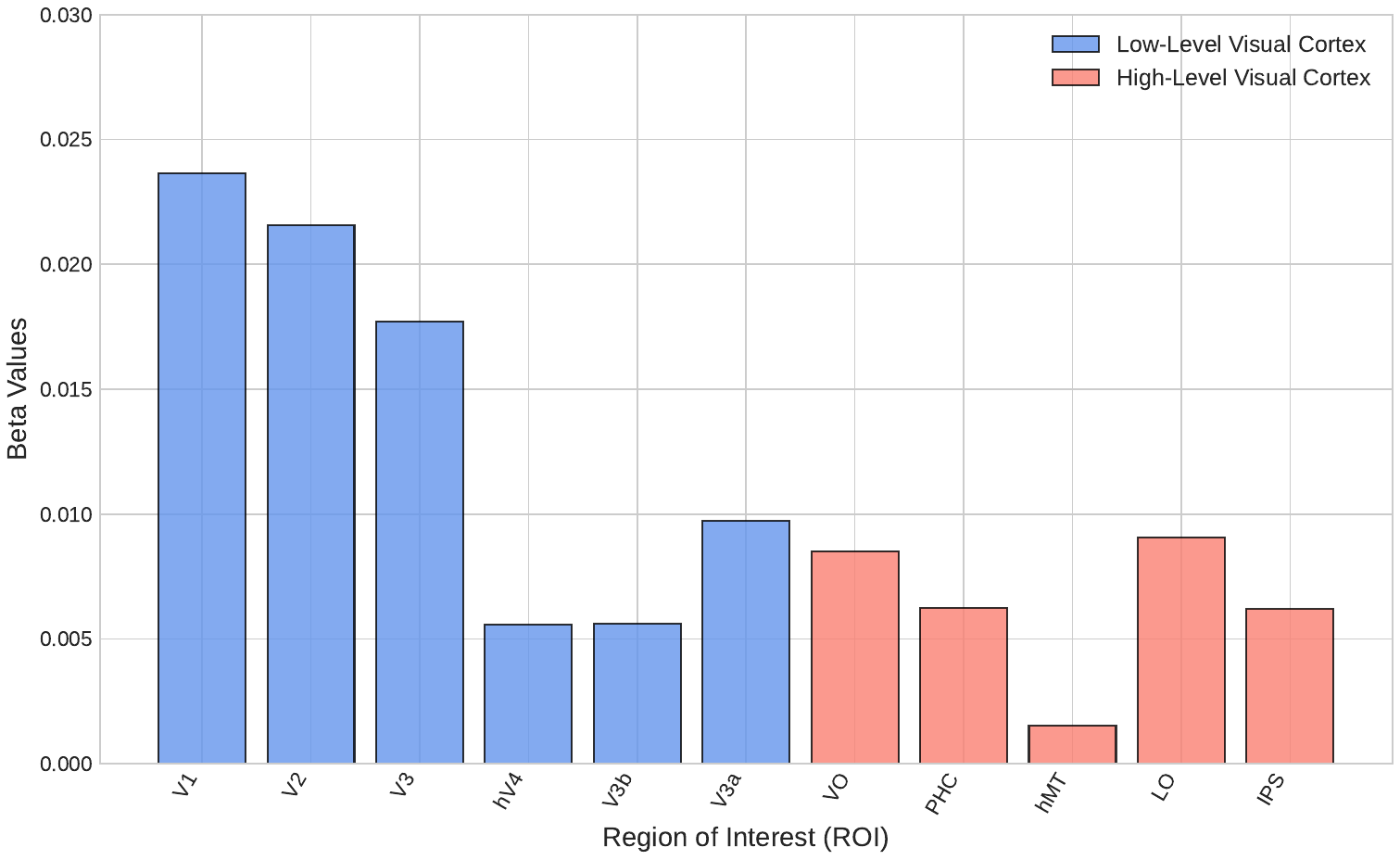}
        \small (B) Image Semantic Beta Value
    \end{minipage}
    \hfill
    \begin{minipage}{0.32\textwidth}
        \centering
        \includegraphics[width=\linewidth]{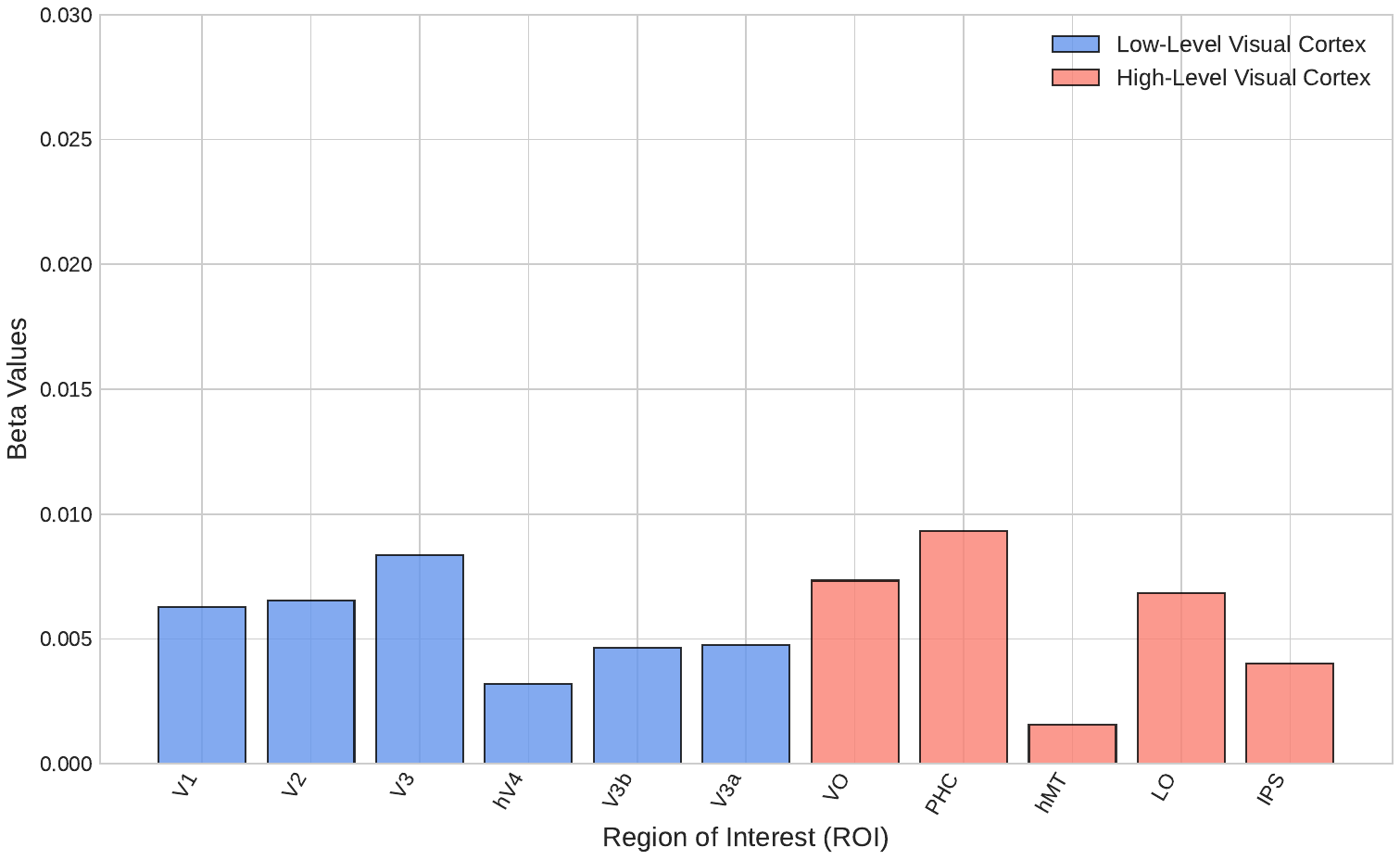}
        \small (C) Text Semantic Beta Value
    \end{minipage}

    \vspace{0.5cm} 

    \begin{minipage}{0.32\textwidth}
        \centering
        \includegraphics[width=\linewidth]{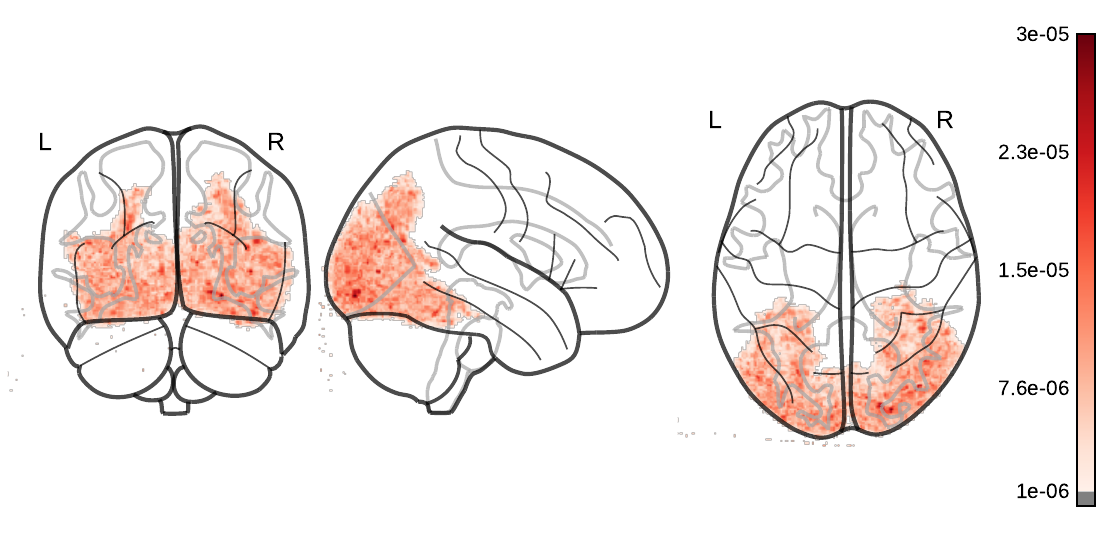}
        \small (D) Image Detail Glass Brain
    \end{minipage}
    \hfill
    \begin{minipage}{0.32\textwidth}
        \centering
        \includegraphics[width=\linewidth]{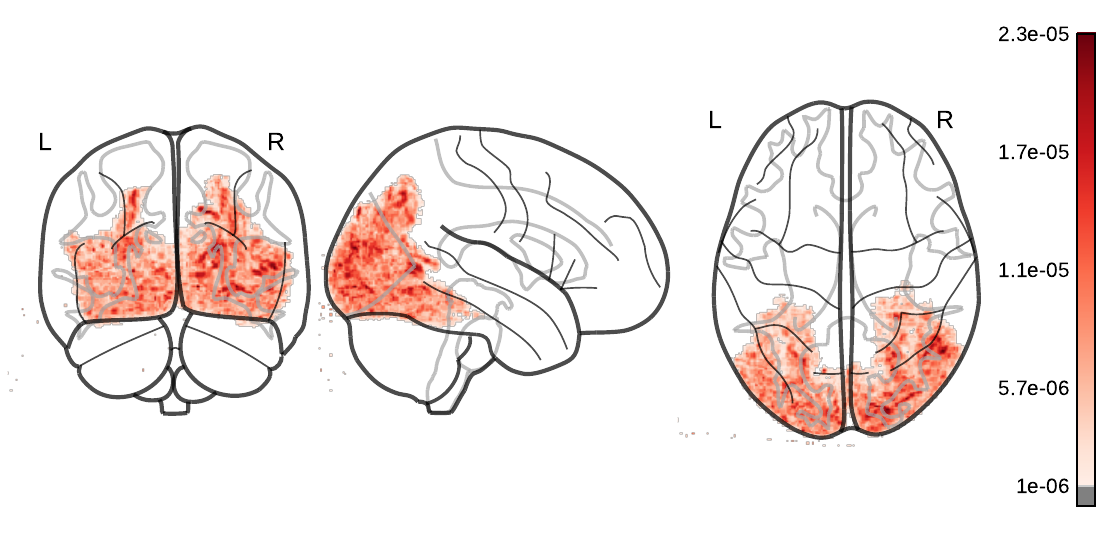}
        \small (E) Image Semantic Glass Brain
    \end{minipage}
    \hfill
    \begin{minipage}{0.32\textwidth}
        \centering
        \includegraphics[width=\linewidth]{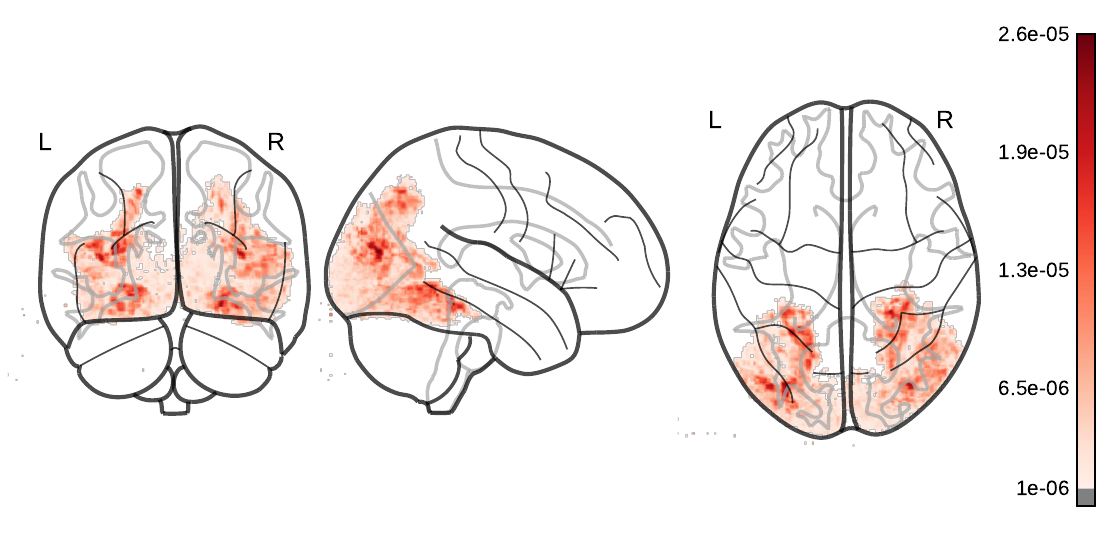}
        \small (F) Text Semantic Glass Brain
    \end{minipage}
    
    \caption{
    The top row (A-C) displays the average Lasso beta weights for various Regions of Interest (ROIs), segregated into low-level (blue) and high-level (red) visual cortex. The bottom row (D-F) visualizes the spatial distribution of these weights on a glass brain.
    }
    \label{fig:appendix_h_projection}
\end{figure*}

To validate the functional alignment of our model's feature branches, we performed a back-projection analysis. Using Lasso regression, we mapped the learned features from the Image Detail, Image Semantic, and Text Semantic branches back to fMRI activity in the `nsdgeneral` ROI. The resulting Lasso weights (beta values), shown for a representative subject in Fig.\ref{fig:appendix_h_projection}, reveal the predictive importance of different brain regions for each feature type.

The analysis confirms our neuro-inspired mapping. As shown in the figure, Image Detail features are most strongly predicted by the low-level visual cortex (e.g., V1-V3). In comparison, the Image Semantic features exhibit a subtle but critical shift in predictive weights: there is a relative decrease in the contribution from low-level visual areas. This pattern is neuroscientifically plausible, as semantic representations are built upon, rather than being independent of, low-level visual features. In stark contrast, Text Semantic features are predominantly associated with high-level ventral stream areas (e.g., PHC, VO, LO), reflecting their abstract nature. This analysis provides strong evidence that BrainMCLIP learns a functionally plausible mapping, successfully differentiating feature representations in accordance with the brain's visual hierarchy.

